\documentclass[runningheads]{llncs}
\usepackage{graphicx}
\usepackage{amsmath,amssymb} % define this before the line numbering.
\usepackage{color}
\usepackage[width=122mm,left=12mm,paperwidth=146mm,height=193mm,top=12mm,paperheight=217mm]{geometry}
\usepackage[compatibility=false]{caption}
\usepackage{subcaption}
\usepackage{makecell}
\usepackage{multirow}
\usepackage[export]{adjustbox}

\begin{document}
% \linenumbers
\pagestyle{headings}
\mainmatter

\title{Deep Markov Random Field for Image Modeling} % Replace with your title

\titlerunning{Deep Markov Random Field for Image Modeling}

\author{Zhirong Wu~~~~~~Dahua Lin~~~~~~Xiaoou Tang}

\authorrunning{Zhirong Wu, Dahua Lin, Xiaoou Tang}
%Please write out author names in full in the paper, i.e. full given and family names. 
%If any authors have names that can be parsed into FirstName LastName in multiple ways, please include the correct parsing, in a comment to the volume editors:
%\index{Lastnames, Firstnames}
%(Do not uncomment it, because you may introduce extra index items if you do that...)

\institute{Department,\\
        University\\
        \email{ \{author1,author2\}@univ.edu}
}
\institute{The Chinese University of Hong Kong~~~~}

\newtheorem{Proposition}{proposition}
\newcommand{\misscite}{\textcolor{red}{[C]}}
\newcommand{\eg}{\textit{e.g.}}
\newcommand{\ie}{\textit{i.e.}}

\newcommand{\argmax}{\mathop{\mathrm{argmax}}}
\newcommand{\argmin}{\mathop{\mathrm{argmin}}}
\newcommand{\Rsp}{\mathbb{R}}

\newcommand{\va}{\mathbf{a}}
\newcommand{\vx}{\mathbf{x}}
\newcommand{\vh}{\mathbf{h}}
\newcommand{\mW}{\mathbf{W}}
\newcommand{\mR}{\mathbf{R}}
\newcommand{\mQ}{\mathbf{Q}}
\newcommand{\vtheta}{\boldsymbol{\theta}}

\newcommand{\hset}{\mathcal{H}}
\newcommand{\Nb}{\mathcal{N}}

\maketitle

\begin{abstract}
Markov Random Fields (MRFs), a formulation widely used in generative image modeling, have long been plagued by the lack of expressive power. 
This issue is primarily due to the fact that conventional MRFs formulations tend to use \emph{simplistic} factors to capture local patterns. 
In this paper, we move beyond such limitations, and propose a novel MRF model that uses fully-connected neurons to express the complex interactions among pixels. 
Through theoretical analysis, we reveal an inherent connection between this model and recurrent neural networks, and thereon derive an approximated feed-forward network that couples multiple RNNs along opposite directions.
This formulation combines the expressive power of deep neural networks and the cyclic dependency structure of MRF in a unified model,  
bringing the modeling capability to a new level. 
The feed-forward approximation also allows it to be efficiently learned from data.
Experimental results on a variety of low-level vision tasks show notable improvement over state-of-the-arts. 
\end{abstract}

\section{Introduction}
\label{sect:intro}

\begin{figure}[t]
\centering
             \includegraphics[width=1\textwidth]{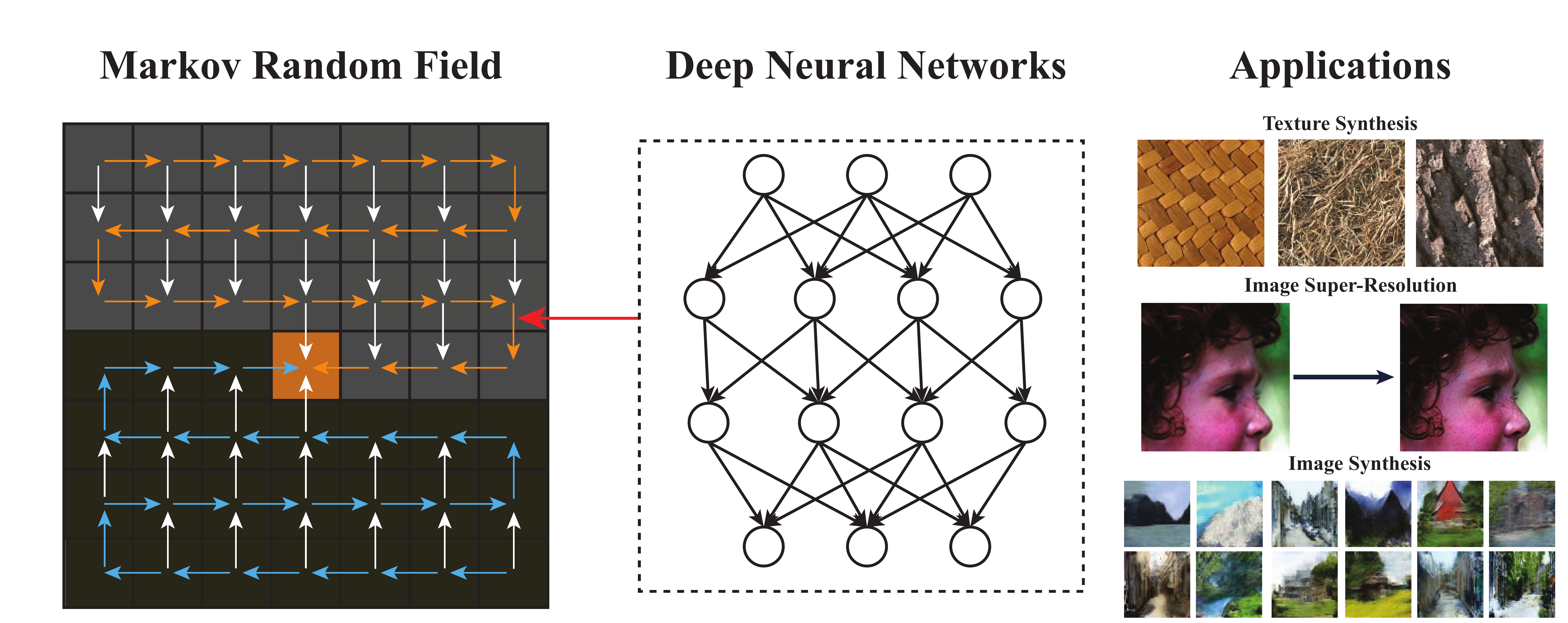}
%              %\caption{}
             %\label{}
%\vspace{-3mm}
\caption{We present a new class of markov random field models whose potential functions are expressed by powerful deep neural networks. We show applications of the model on texture synthesis, image super-resolution and image synthesis.}
%\vspace{-2mm}
\label{fig:teaser}
\end{figure}

%% BACKGROUND: Why this is an important problem
Generative image models play a crucial role in a variety of image processing and computer vision tasks,
such as denoising~\cite{portilla2003image}, super-resolution~\cite{freeman2000learning}, inpainting~\cite{bertalmio2000image}, and image-based rendering~\cite{mcmillan1995plenoptic}. 
As repeatedly shown by previous work~\cite{huang1999statistics}, 
the success of image modeling, to a large extent, hinges on whether the model can successfully capture the spatial relations among pixels. 

%% TWO CATEGORIES
Existing image models can be roughly categorized as \emph{global models} and \emph{low-level models}.
Global models~\cite{turk1991face,wright2010sparse,hinton2002training}
usually rely on compressed representations to capture the global structures. 
Such models are typically used for describing objects with regular structures, \eg~faces. 
For generic images, low-level models are more popular.
Thanks to their focus on local patterns instead of global appearance, 
low-level models tend to generalize much better, especially when there can be vast variations in the image content.

%% Discussing the limitations of MRF 
Over the past decades, \emph{Markov Random Fields (MRFs)} have evolved into one of the most popular models for low-level vision. Specifically, the clique-based structure makes them particularly well suited for capturing local relations among pixels. 
Whereas MRFs as a generic mathematical framework are very flexible and provide immense expressive power, 
the performance of many MRF-based methods still leaves a lot to be desired when faced with challenging conditions. 
This occurs due to the widespread use of \emph{simplistic} potential functions that largely restrict the expressive power of MRFs. 

%% Deep neural networks
In recent years, the rise of \emph{Deep Neural Networks (DNN)} has profoundly reshaped the landscape of many areas in computer vision. 
The success of DNNs is primarily attributed to its unparalleled expressive power, particularly their strong capability of modeling complex variations. 
However, DNNs in computer vision are mostly formulated as end-to-end convolutional networks (CNN) for classification or regression. The modeling of local interactions among pixels, which is crucial for many low-level vision tasks, has not been sufficiently explored.

%% Our model: MRF + DNN
The respective strengths of MRFs and DNNs inspire us to explore a new approach to low-level image modeling, that is, to bring the expressive power of DNNs to an MRF formulation.  
Specifically, we propose a generative image model comprised of a grid of \emph{hidden states}, each corresponding to a pixel. These latent states are connected to their neighbors -- together they form an MRF. 
Unlike in classical MRF formulations, we use fully connected layers to express the relationship among these variables, thus substantially improving the model's ability to capture complex patterns. 

%% RNN related
Through theoretical analysis, we reveal an inherent connection between our MRF formulation and the RNN~\cite{werbos1990backpropagation}, 
which opens an alternative way to MRF formulation. 
However, they still differ fundamentally: the dependency structure of an RNN is \emph{acyclic}, while that of an MRF is \emph{cyclic}. 
Consequently, the hidden states cannot be inferred in a single \emph{feed-forward} manner as in a RNN. 
This posts a significant challenge -- how can one derive the back-propagation procedure without a well-defined forward function?

Our strategy to tackle this difficulty is to \emph{unroll an iterative inference procedure into a feed-forward function}.
This is motivated by the observation that while the inference is iterative, each cycle of updates is still a feed-forward procedure. 
Following a carefully devised scheduling policy, which we call the \emph{Coupled Acyclic Passes (CAP)}, the inference can be unrolled into multiple RNNs operating along opposite directions that are coupled together. 
In this way, local information can be effectively propagated over the entire network, where each hidden state can have a complete picture of its context from all directions. 

%% Experiment & summary

The primary contribution of this work is a new generative model that unifies MRFs and DNNs in a novel way, as well as a new learning strategy that makes it possible to learn such a model using mainstream deep learning frameworks. 
It is worth noting that the proposed method is generic and can be adapted to a various problems. 
In this work, we test it on a variety of low-level vision tasks, including texture synthesis, image super-resolution, and image synthesis. 

% Organization of the rest of the paper

\section{Related Works}
\label{sect:related}

In this paper, we develop a generative image model that incorporates the expressive power of deep neural networks with an MRF. 
This work is related to several streams of research efforts, but moves beyond their respective limitations. \\[-8pt]

%\vspace{3pt}

\noindent
\textbf{Generative image models.}
%\paragraph{\bf Generative image models.}
Generative image models generally fall into two categories: parametric models and non-parametric models. 
\emph{Parametric models} typically use a compressed representation to capture an image's global appearance. 
In recent years, deep networks such as autoencoders~\cite{kingma2013auto} and adversarial networks~\cite{goodfellow2014generative,denton2015deep} have achieved substantial improvement in generating images with regular structures such as faces or digits. 
%However, such models are still not capable of generic images modeling. 
%Attempts of using these networks in natural images usually leads to results that are overly noisy or blurry.
%
\emph{Non-parametric models}, 
including \emph{pixel-based sampling}~\cite{efros1999texture,wei2000fast,hertzmann2001image} and 
\emph{patch-based sampling}~\cite{efros2001image,hays2007scene,lalonde2007photo},
instead rely on a large set of exemplars to capture local patterns. 
Whereas these methods can produce high quality images with local patterns directly sampled from realistic images.
Exhaustive search over a large exemplar set limits their scalability and often leads to computational difficulties.
Our work draws inspiration from both lines of work. 
By using DNNs to express local interactions in an MRF, our model can capture highly complex patterns while maintaining strong scalability. \\[-8pt]

%\vspace{3pt}

\noindent
\textbf{Markov random fields.}
%\paragraph{\bf Markov random fields.}
For decades, MRFs have been widely used for low-level vision tasks, 
including texture synthesis~\cite{cross1983markov}, segmentation~\cite{boykov2001interactive,he2004multiscale}, denoising~\cite{portilla2003image}, and super-resolution~\cite{freeman2000learning}. 
Classical MRF models in earlier work~\cite{geman1984stochastic} use simple hand-crafted potentials (\eg,~Ising models~\cite{ising1925beitrag}, Gaussian MRFs~\cite{rue2005gaussian}) to link neighboring pixels. 
Later, more flexible models such as FRAME~\cite{zhu1998filters} and Fields of Experts~\cite{roth2005fields} were proposed, 
which allow the potential functions to be learned from data. 
However, in these methods, the potential functions are usually parameterized as a set of linear filters, and therefore their expressive power remains very limited. \\[-8pt]

%\vspace{3pt}

\noindent
\textbf{Recurrent neural networks.}
%\paragraph{\bf Recurrent neural networks.}
\emph{Recurrent neural networks (RNNs)}, a special family of deep models, use a chain of nonlinear units to capture sequential relations. 
%RNNs have been shown to be very effective in natural language processing~\cite{mikolov2010recurrent}. 
In computer vision, RNNs are primarily used to model sequential changes in videos~\cite{donahue2015long}, 
visual attention~\cite{mnih2014recurrent,gregor2015draw}, and hand-written digit recognition~\cite{graves2009offline}.
Previous work explores multi-dimensional RNNs~\cite{mdrnn} for scene labeling~\cite{byeon2015scene} as well as object detections~\cite{bell2015inside}.
The most related work is perhaps the use of 2D RNNs for generating gray-scale textures~\cite{theis2015generative} or color images~\cite{oord2016pixel}.
A key distinction of these models from ours is that 2D RNNs rely on an \emph{acyclic graphs} to model spatial dependency, \eg~each pixel depends only on its left and upper neighbors -- this severely limits the spatial coherence. Our model, instead, allows dependencies from all directions via iterative inference unrolling. \\[-8pt]
%It is also worth noting that \emph{Pixel RNN}~\cite{oord2016pixel} can only work with a fixed image size, \eg~ $32 \times 32$, while ours can work with arbitrary sizes. 

%\vspace{3pt}

\noindent
\textbf{MRF and neural networks.}
Connections between both models have been discussed long ago~\cite{rangarajan1991markov}. With the rise of deep learning, recent work on image segmentation~\cite{zheng2015conditional,chen2014semantic} uses mean field method to approximate a conditional random field (CRF) with CNN layers. 
A hybrid model of CNN and MRF has also been proposed for human pose estimation~\cite{tompson2014joint}. 
These works primarily target prediction problems (\eg~segmentation) and are not as effective at capturing complex pixel patterns in a purely generative way.

\section{Deep Markov Random Field}
%We propose to embed RNNs into markov random field $G = (V, E)$ to capture the complex spatial dependencies. The vertexes $V$ correspond to the hidden states of the RNN; the edges $E$ correspond to the recurrent connections. While we can actually cast any graph structures into the formulation of RNNs, we focus on 4-neighborhood markov random field for modeling images.

\begin{figure}[t]
\centering
     \begin{subfigure}{0.55\textwidth}
			\centering
             \includegraphics[width=1\textwidth]{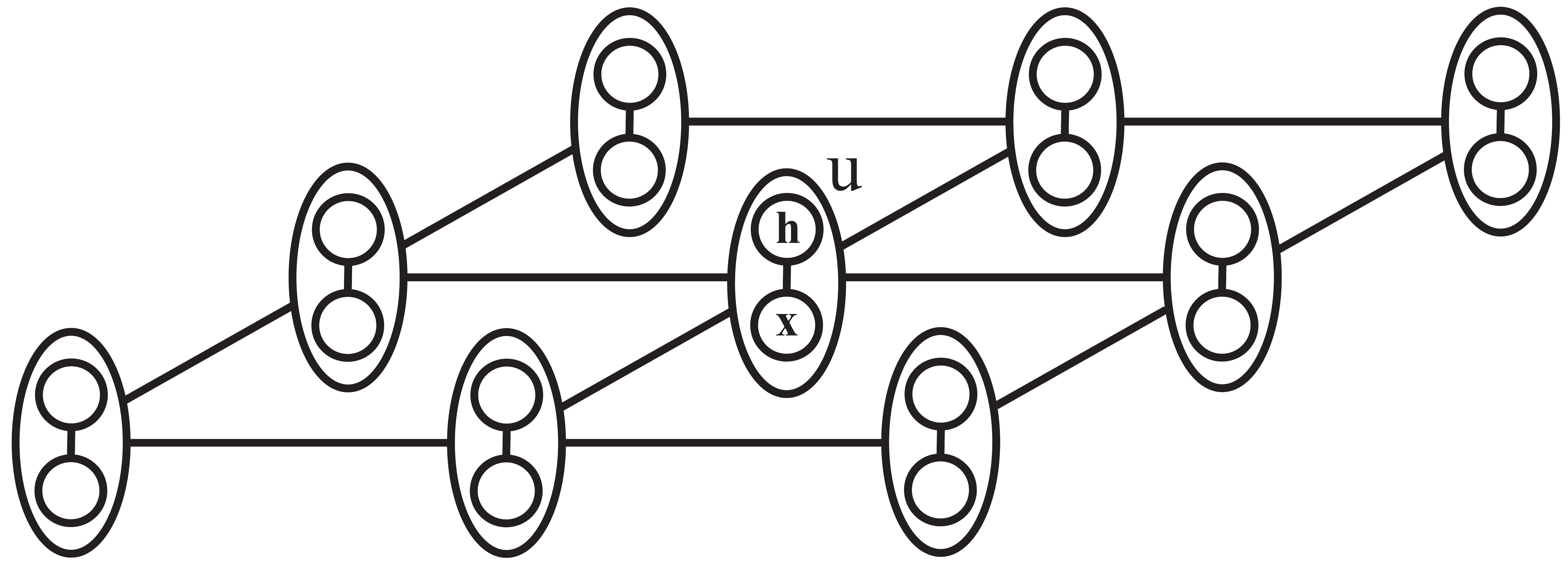}
              %\caption{}
             %\label{}
     \end{subfigure}
     ~
     	\begin{subfigure}{0.32\textwidth}
			\centering
             \includegraphics[width=1\textwidth]{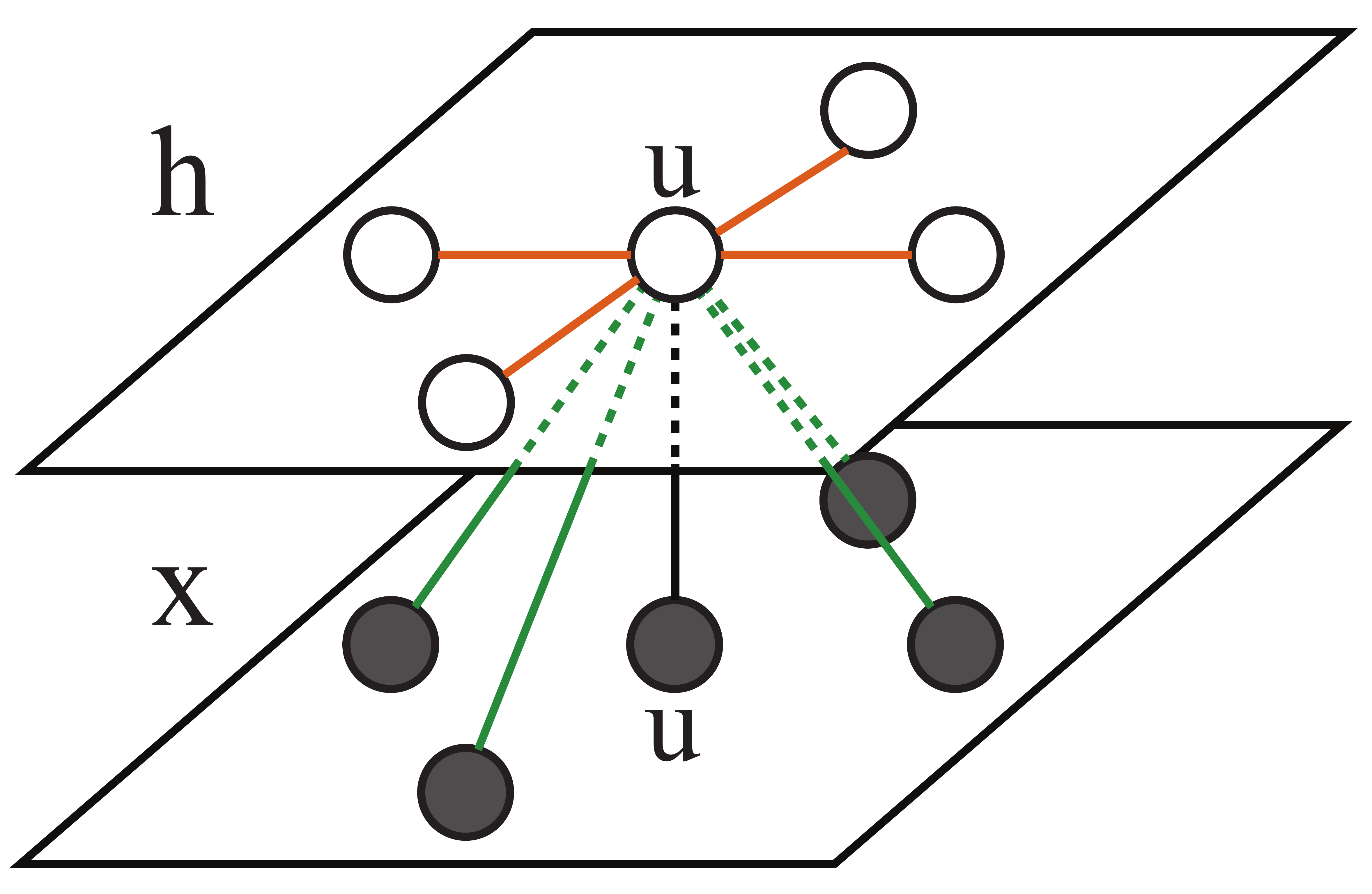}
              %\caption{}
              %\label{}
     \end{subfigure}
%\vspace{-3mm}
\caption{Graphical model of deep MRFs. {\bf Left}: The hidden states and the pixels together form an MRF. {\bf Right}: Each hidden state connects to the neighboring states, the neighboring pixels, and the pixel at the same location.}
%\vspace{-2mm}
\label{fig:pgm}
\end{figure}

%We start from the formulation of MRF, and see how it can be casted into graph-based RNNs.
The primary goal of this work is to develop a generative model for images that can express complex local relationships among pixels while being tractable for inference and learning. 
Formally, we consider an image, denoted by $\vx$, as an \emph{undirected graph} with a grid structure, as shown in Figure~\ref{fig:pgm} left. 
Each node $u$ corresponds to a pixel $x_u$. 
To capture the interactions among pixels, we introduce, $h_u$, a hidden variable for each pixel denoting the hidden state corresponding to the pixel $x_u$. 
In the graph, each node $u$ has a neighborhood, denoted by $\Nb_u$. Particularly, we use the \emph{4-connected neighborhood} of a 2D grid in this work.

\paragraph{\bf Joint Distribution.}
We consider three kinds of dependencies: 
(1) the dependency between a pixel $x_u$ and its corresponding hidden state $h_u$,
(2) the dependency between a hidden state $h_u$ and a neighbor $h_v$ with $v \in \Nb_u$, and 
(3) the dependency between a hidden state $h_u$ and a neighboring pixel $x_v$. 
They are respectively captured by factors 
$\zeta(x_u, h_u)$, $\phi(h_u, h_v)$, and $\psi(h_u, x_v)$. 
In addition, we introduce a regularization factor $\lambda(h_u)$ for each hidden state, 
which gives us the leeway to encourage certain distribution over the state values. 
Bringing these factors together, we formulate an MRF to express the joint distribution:
\begin{equation} \label{eq:mrf}
	p(\vx, \vh) = \frac{1}{Z} 
	\prod_{u \in V} \zeta(x_u, h_u) 
	\prod_{(u, v) \in E} \left( \phi(h_u, h_v) \psi(h_u, x_v) \psi(h_v, x_u) \right)
	\prod_{u \in V} \lambda(h_u).
\end{equation}
Here, $V$ and $E$ are respectively the set of vertices and that of the edges in the image graph, $Z$ is a normalizing constant. 
Figure~\ref{fig:pgm} shows it structure. 

\paragraph{\bf Choices of Factors.}
Whereas the MRF provides a principled way to express the dependency structure, 
the expressive power of the model still largely depends on the specific forms of the factors that we choose.
For example, the modeling capacity of classical MRF models are limited by their simplistic factors. 

Below, we discuss the factors that we choose for the proposed model. 
First, the factor $\zeta(x_u, h_u)$ determines how the pixel values are generated from the hidden states. 
Considering the stochastic nature of natural images, we formalize this generative process as a \emph{Gaussian mixture model (GMM)}.
The rationale behind is that pixel values are on a low-dimensional space, 
where a GMM with a small number of components can usually provide a good approximation to an empirical distribution. 
Specifically, we fix the number of components to be $K$, and 
consider the concatenation of component parameters as the linear transform of the hidden state, $h_u^T\mathbf{Q} = ((\pi_u^c, \mu_u^c, \Sigma_u^c))_{c=1}^K$, where $\mathbf{Q}$ is a weight matrix of model parameters. 
In this way, the factor $\zeta(x_u, h_u)$ can be written as
\begin{equation} \label{eq:gmm}
	\zeta(x_u, h_u) \triangleq p_{\mathrm{GMM}}(x_u | h_u) = \sum_{c=1}^{K}\pi_u^c N(x_u|\mu_u^c, \Sigma_u^c).
\end{equation}
To capture the rich interactions among pixels and their neighbors, we formulate the relational factors $\phi(h_u, h_v)$ and $\psi(h_u, x_v)$ 
with \emph{fully connected} forms:
\begin{equation}
\phi(h_u, h_v) = \exp\left( h_u^T \mW h_v \right), \quad
\psi(h_u, x_v) =  \exp\left(h_u^T \mR x_v\right).
\end{equation}
Finally, to control the value distribution of the hidden states, we further incorporate a regularization term over $h_u$, as 
\begin{equation}
\lambda(h_u) 
= \exp \left( - \mathbf{1}^T \eta(h_u) \right) 
= \exp \left( -\eta(h_u^{(1)}) - \cdots - \eta(h_u^{(d)}) \right).
\end{equation} 
Here, $\eta$ is an element-wise nonlinear function and $d$ is the dimension of $h_u$.
%
% Discussion
In summary, 
the use of GMM in $\zeta(x_u, h_u)$ effectively accounts for the variations in pixel generation, 
the fully-connected factors $\phi(h_u, h_v)$ and $\psi(h_u, x_v)$ enable the modeling of complex interactions among neighbors,
while the regularization term $\lambda(h_u)$ provides a way to explicitly control the distribution of hidden states. 
Together, they substantially increase the capacity of the MRF model.

\begin{figure}[t]
\begin{tabular}{c|cccc}
\footnotesize
\centering
     \begin{subfigure}{0.20\textwidth}
			\centering
             \includegraphics[width=1\textwidth]{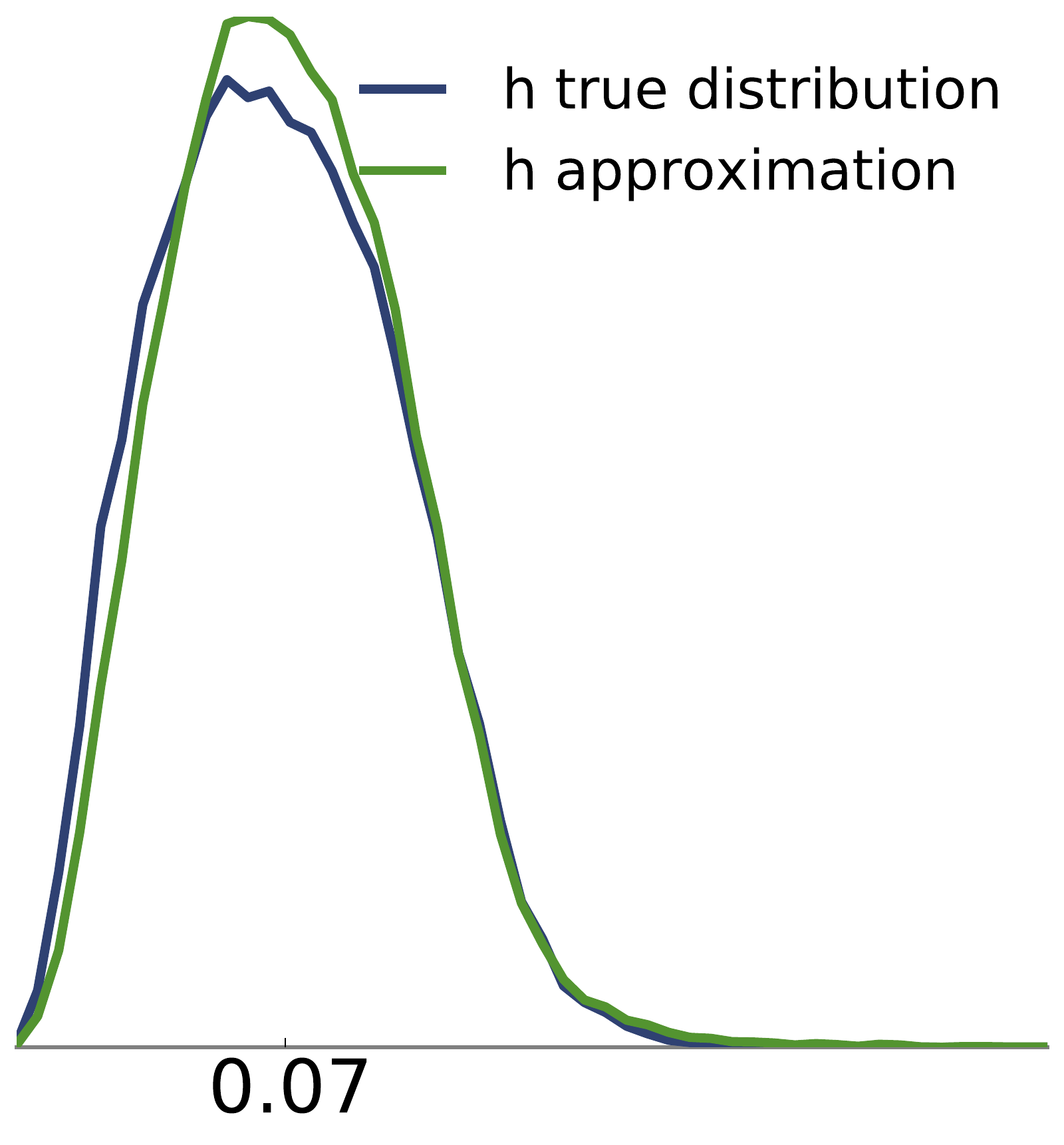}
              %\caption{}
             %\label{}
     \end{subfigure}
                    ~ & ~
	\begin{subfigure}{0.18\textwidth}
			\centering
             \includegraphics[width=1\textwidth]{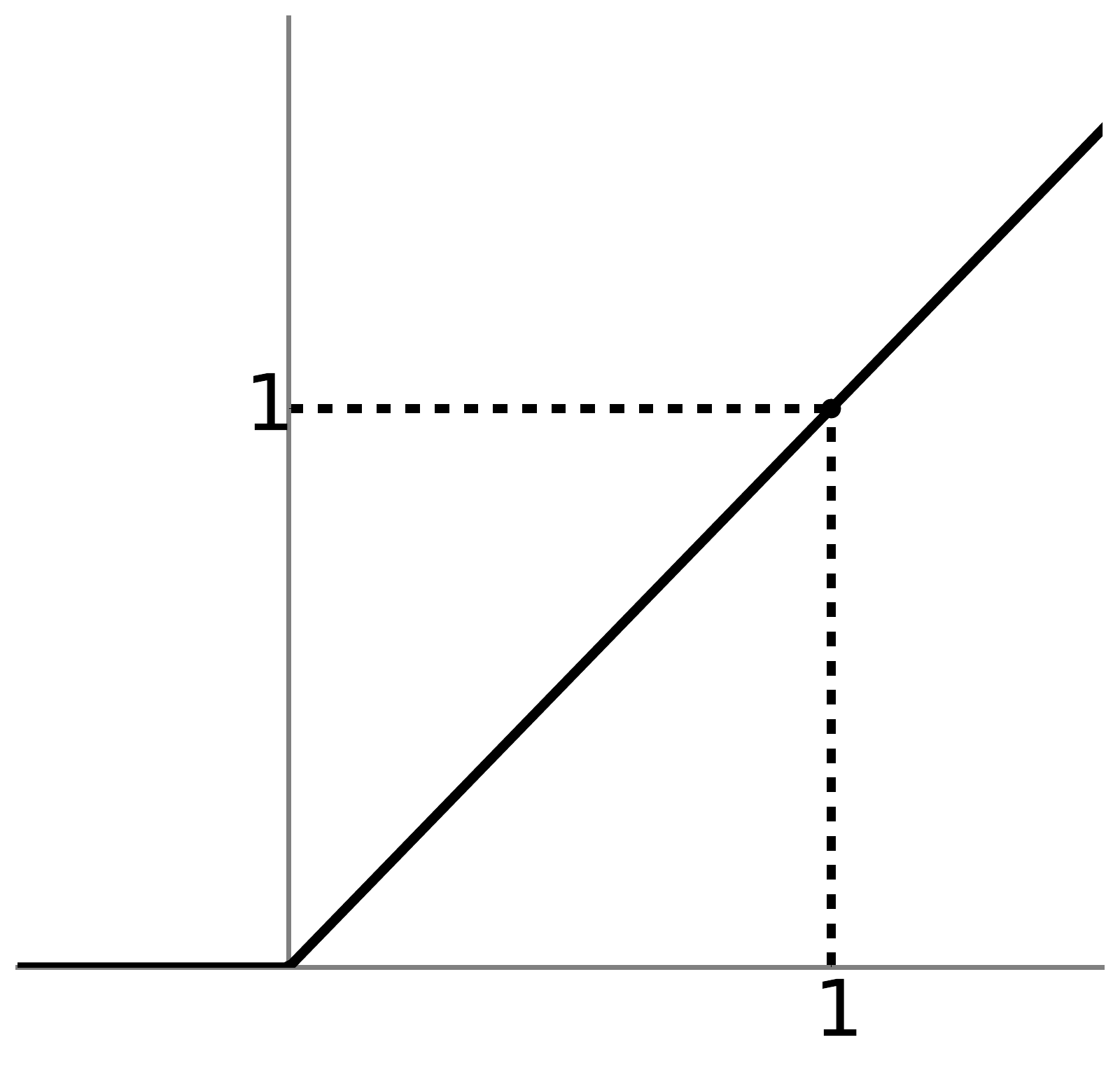}
              %\caption{}
              %\label{}
     \end{subfigure}
      &
     \begin{subfigure}{0.18\textwidth}
			\centering
             \includegraphics[width=1\textwidth]{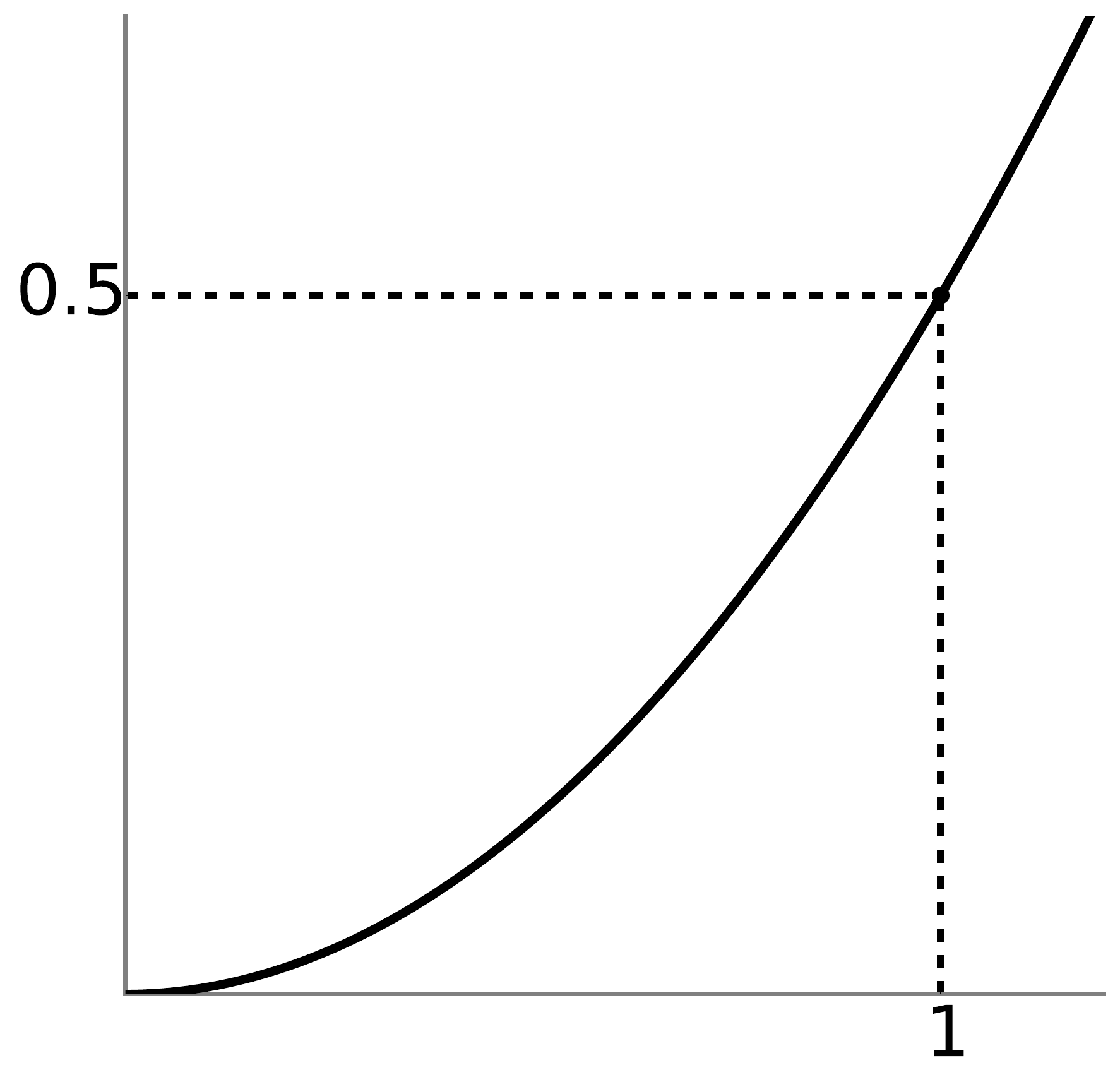}
              %\caption{}
             %\label{}
     \end{subfigure}
           & 
     \begin{subfigure}{0.18\textwidth}
			\centering
             \includegraphics[width=1\textwidth]{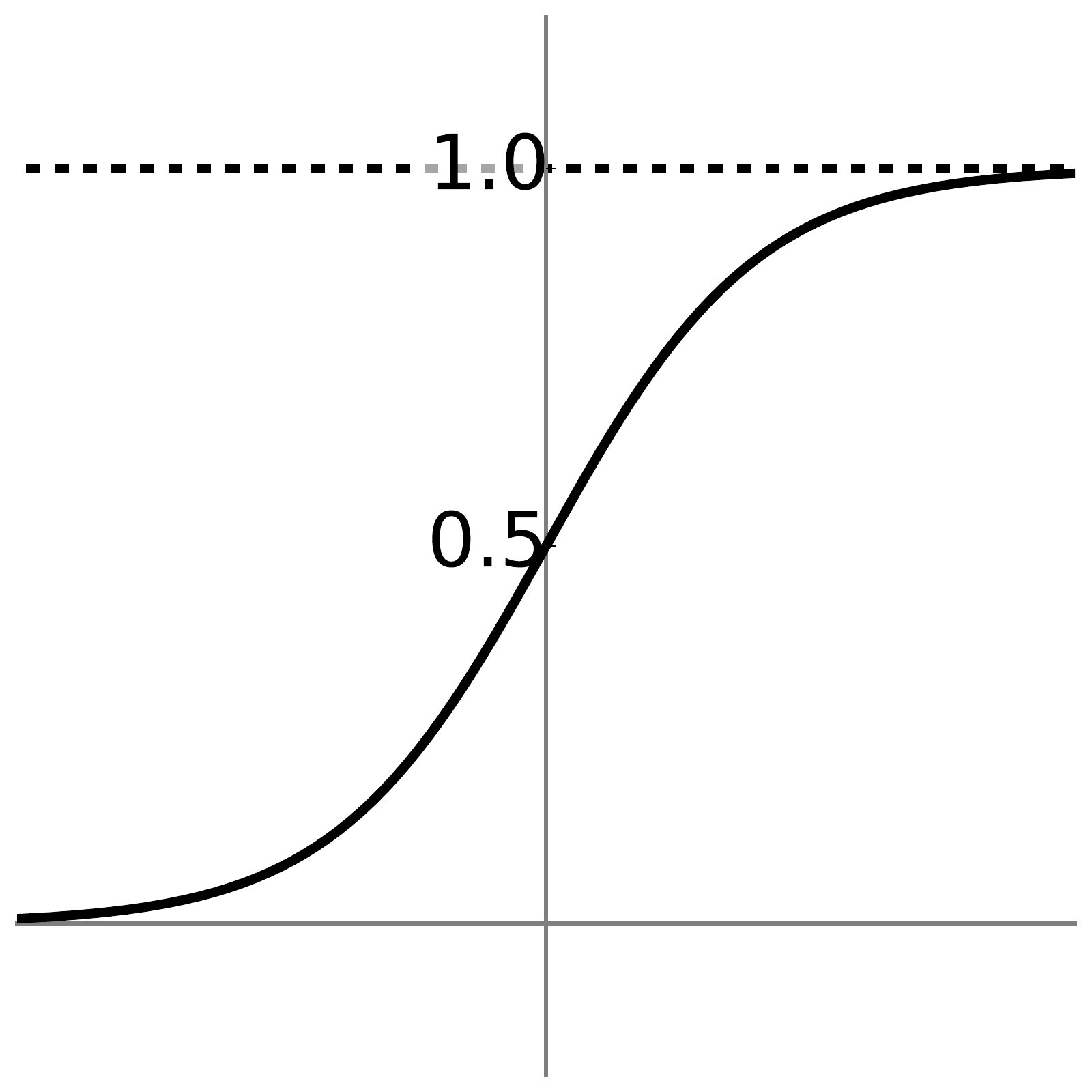}
              %\caption{}
             %\label{}
     \end{subfigure}
           &
     \begin{subfigure}{0.18\textwidth}
			\centering
             \includegraphics[width=1\textwidth]{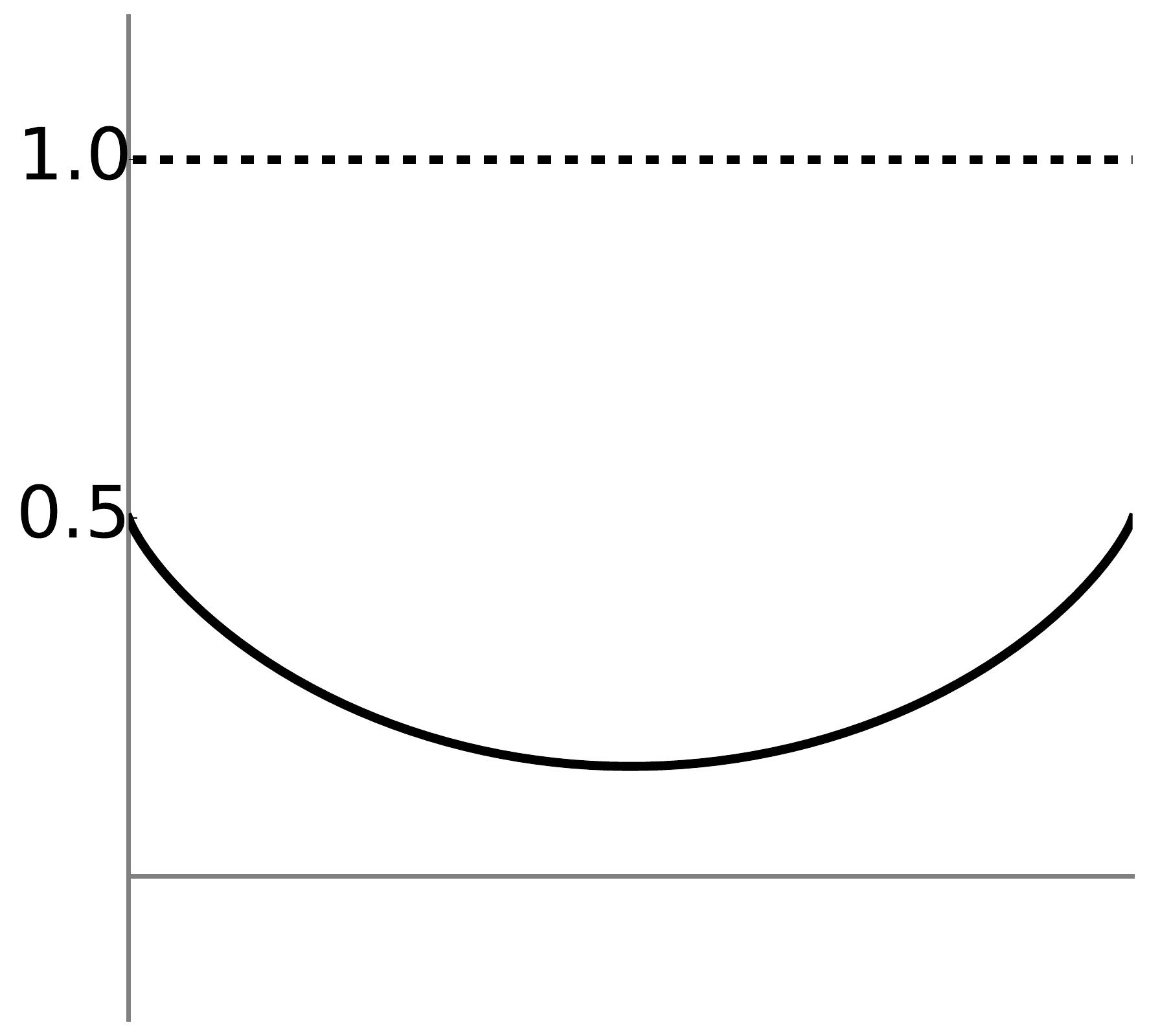}
              %\caption{}
             %\label{}
     \end{subfigure}
      \\
     $\mathbf{h}$ approximation & ReLU & Regularizer & Sigmoid & Regularizer  \\
\end{tabular}
%\vspace{-3mm}
\caption{ {\bf Left} shows the numerical simulation of approximated inference for the hidden variables. {\bf Right} shows the ReLU, sigmoid activation function and their corresponding regularizations for the hidden variables.}
%\vspace{-2mm}
\label{fig:reg}
\end{figure}

\paragraph{\bf Inference of Hidden States.}
With this MRF formulation, the posterior distribution of the hidden state $h_u$, conditioned on all other variables, 
is given by
\begin{equation} \label{eq:cond_distr}
	p \left(h_u \mid x_u, x_{\Nb_u}, h_{\Nb_u} \right)
	\propto 
	\zeta(x_u, h_u) \lambda(h_u) \cdot 
	\prod_{v \in \Nb_u} \phi(h_u, h_v) \psi(h_u, x_v).
\end{equation}
Here, $h_u$ depends on its neighboring states, the corresponding pixel values, as well as that of its neighbors.
Since the pixel $x_u$ and its neighboring pixels $x_{\Nb_u}$ are highly correlated, to simplify our later computations, we approximate the posterior distribution as,
%For natural images, neighboring pixels are highly correlated, or in other words, the information conveyed by adjacent pixels are highly redundant. We therefore hypothesize that the posterior distribution of $h_u$ can be approximated as
\begin{equation} \label{eq:h_approx}
	p \left(h_u \mid x_u, x_{\Nb_u}, h_{\Nb_u} \right) 
	\simeq
	p \left(h_u \mid x_{\Nb_u}, h_{\Nb_u} \right)
	\propto \lambda(h) \prod_{v \in \Nb_u} \phi(h, h_v) \psi(h, x_v).
\end{equation}
%
%To verify this hypothesis, we performed extensive simulations, and found that the posterior distribution given by
%Eq.\eqref{eq:cond_distr} and the approximation given by Eq.\eqref{eq:h_approx} are indeed very close to each other, 
%as illustrated in Figure~\textcolor{red}{[R]}.
%
We performed numerical simulations for this approximation. They are indeed very close to each other,  as illustrated in Figure~\ref{fig:reg}. Consequently, the MAP estimate of $h_u$ can be \emph{approximately} computed from its neighbors. It turns out that this optimization problem has an analytic solution given by,
%{\small
%\begin{equation} \label{eq:h_rnn}
%	\tilde{h}_u 
%	= \argmax_h \left\{ h^T \left( \sum_{v \in \Nb_u} \mW h_v + \mR %x_v \right) - \mathbf{1}^T \eta(h) \right\}.
%\end{equation}
%}
%Taking the derivatives to zeros, it turns out that this optimization problem has an analytic solution given by, 
{\small
\begin{equation} \label{eq:h_rnn}
	\tilde{h}_u 
	= \sigma \left( \sum_{v \in \Nb_u} \mW h_v + \mR x_v \right).
\end{equation}
}
Here, $\sigma$ is an element-wise function that is related to $\eta$ as $\sigma^{-1}(z) = \eta'(z)$,
where $\eta'$ is the first-order derivative \textit{w.r.t.}~$\eta$, and $\sigma^{-1}$ the inverse function of $\sigma$. 

\paragraph{\bf Connections to RNNs.}

We observe that Eq.\eqref{eq:h_rnn} has a form that is similar to the feed-forward computations in 
\emph{Recurrent Neural Networks (RNN)}~\cite{werbos1990backpropagation}.
In this sense, we can view the feed-forward RNN as an MAP inference process for MRF models. 
Particularly, given the RNN computations in the form of Eq.\eqref{eq:h_rnn}, 
one can formulate an MRF as in Eq.\eqref{eq:mrf}, where
regularization function $\eta$ can be derived from $\sigma$ according to the relation $\sigma^{-1}(z) = \eta'(z)$, as
\begin{equation}
	\eta(h) = \int_b^h \sigma^{-1}(z) dz + C.
\end{equation}
Here, $b$ is the minimum of the domain of $h$, which can be $-\infty$, and $C$ is an arbitrary constant. 
This connection provides an alternative way to formulate an MRF model.
More importantly, in this way, RNN models that have been proven to be successful can be readily transferred to an MRF formulation.  
Figure~\ref{fig:reg} shows the regularization functions $\eta(h)$ corresponding to popular activation functions in RNNs, 
such as \emph{sigmoid} and \emph{ReLU}~\cite{krizhevsky2012imagenet}.

%\begin{figure}[t]
%\centering
%	\begin{subfigure}{0.45\textwidth}
%			\centering
%             \includegraphics[width=1\textwidth]{figures/2doutline.pdf}
%              %\caption{}
%              %\label{}
%     \end{subfigure}
%     ~
%     \begin{subfigure}{0.30\textwidth}
%			\centering
%             \includegraphics[width=1\textwidth]{figures/2dfield.pdf}
%              %\caption{}
%             %\label{}
%     \end{subfigure}
%%\vspace{-3mm}
%\caption{2D recurrent neural networks. Left: 2D RNN scans the image in a raster scan order. Right: Each state depends on the upper left corner, and blinds on the upper right corner. This may cause serious spatial discrepancies.}
%%\vspace{-2mm}
%\label{fig:2drnn}
%\end{figure}

\section{Learning via Coupled Recurrent Networks}

Except for special cases~\cite{pearl2014probabilistic}, inference and learning on MRFs is generally intractable.
Conventional estimation methods~\cite{li2009markov,hinton2002training,salakhutdinov2009learning} 
either take overly long time to train or tend to yield poor estimates, especially for models with a high-dimensional parameter space. 
In this work, we consider an alternative approach to MRF learning, 
which allows us to draw on deep learning techniques~\cite{duchi2011adaptive,graves2013generating} 
that have been proven to be highly effective~\cite{krizhevsky2012imagenet}.

\paragraph{\bf Variational Learning Principle.}

Estimation of probabilistic models based on the \emph{maximum likelihood} principle is often intractable when
the model contains hidden variables. 
\emph{Expectation-maximization}~\cite{em1977} is one of the most widely used ways to tackle this problem, 
which iteratively calculates the posterior distribution of $\vh_i$ (in E-steps) and then optimizes $\vtheta$ (in M-steps) as
\begin{equation}
	\hat{\vtheta} = \argmax_{\vtheta} \ \frac{1}{n} \sum_{i=1}^n 
	\mathrm{E}_{p(\vh_i | \vx_i, \vtheta)} \left\{ \log p(\vx_i, \vh_i | \vtheta) \right\}.
\end{equation}
Here, $\vtheta=\{\mW, \mQ, \mR\}$ is the model parameter, $\vx_i$ is the $i$-th image, and $\vh_i$ is the corresponding hidden state. 
As exact computation of this posterior expectation is intractable, 
we approximate it based on $\tilde{\vh}_i$, the MAP estimate of $\vh_i$, as below:
\begin{equation} \label{eq:objective}
	\hat{\vtheta} = \argmax_{\vtheta} \ \frac{1}{n} \sum_{i=1}^n \log p(\vx_i | \tilde{\vh}_i, \vtheta), 
	\text{ with }
	\tilde{\vh}_i \triangleq f(\vx_i, \vtheta).
\end{equation}
This is the \emph{learning objective} of our model.
Here, $f$ is the function that \emph{approximately} infers the latent state $\tilde{\vh}_i$ given an observed image $\vx_i$. 
When the posterior distribution $p(\vh_i | \vx_i, \vtheta)$ is highly concentrated, which is often the case in vision tasks, 
this is a good approximation.
% We note that similar approximation strategies have also been explored in other contexts~\cite{bishop2007gvsd}.
%
For an image $\vx$, $\log p(\vx | \tilde{\vh}, \vtheta)$ can be further expanded as a sum of terms defined on individual pixels:
\begin{equation}
	\log p(\vx | \tilde{\vh}, \vtheta) = \sum_u \log p_{\mathrm{GMM}}(x_u|\tilde{\vh})
	= \sum_u \log \sum_{c=1}^{K}\pi_u^c N(x_u|\tilde{\mu_u^c},\Sigma_u^c),
\end{equation}
where $\tilde{\mu_u^c}=\mu_u^c+\Sigma_u^c(\sum_v \vh_v^T)\mR$.
For our problem, this learning principle can be interpreted in terms of encoding/decoding
-- the hidden states $\tilde{\vh} = f(\vx, \vtheta)$ can be understood as 
an representation that encodes the observed patterns in an image $\vx$,
while $\log p(\vx | \tilde{\vh}, \vtheta)$ measures how well $\tilde{\vh}$ explains the observations.

\begin{figure}[t]
\centering
	\begin{subfigure}{0.45\textwidth}
			\centering
             \includegraphics[width=1\textwidth]{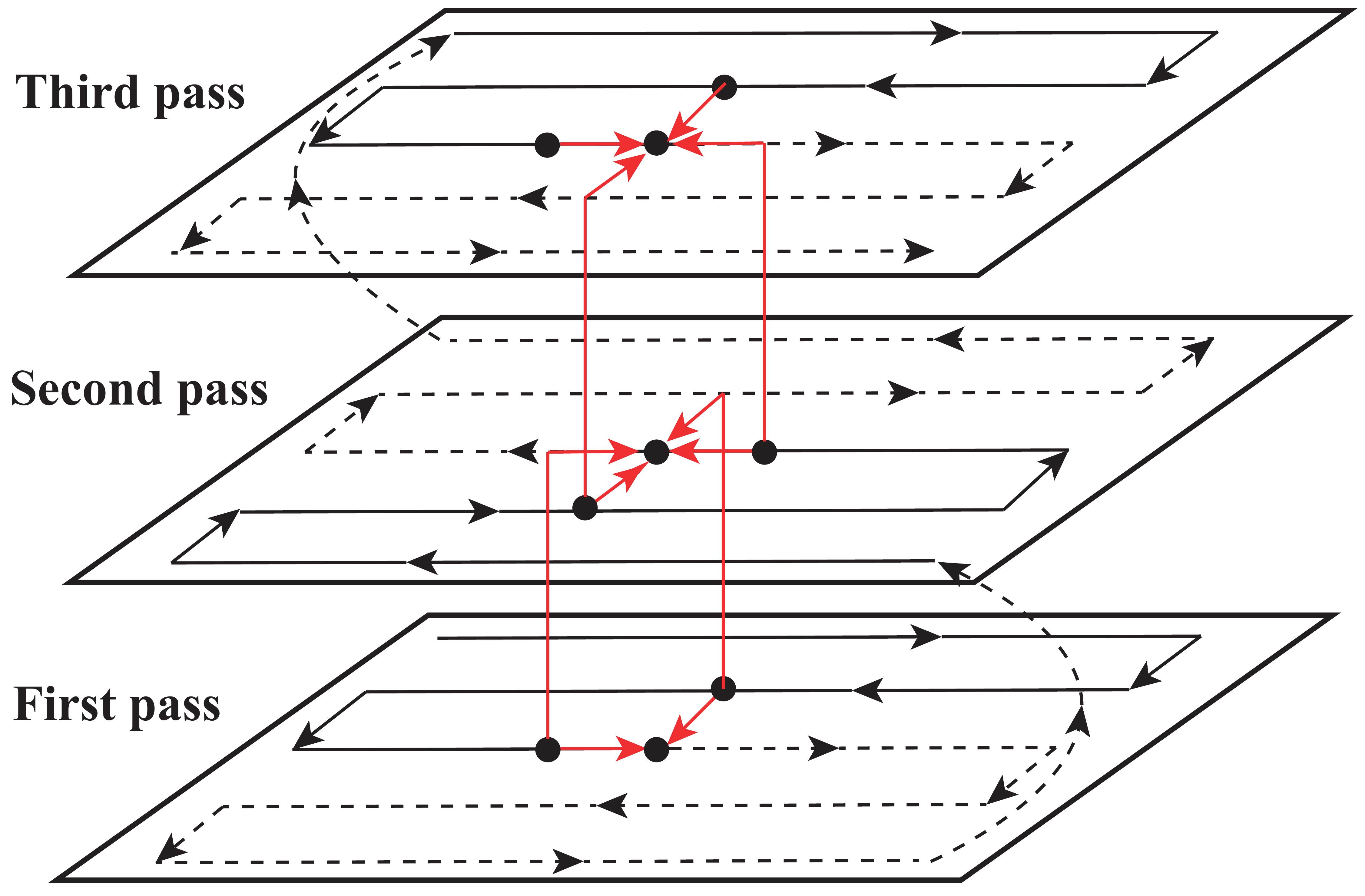}
              %\caption{}
              %\label{}
     \end{subfigure}
     ~
     \begin{subfigure}{0.50\textwidth}
			\centering
             \includegraphics[width=1\textwidth]{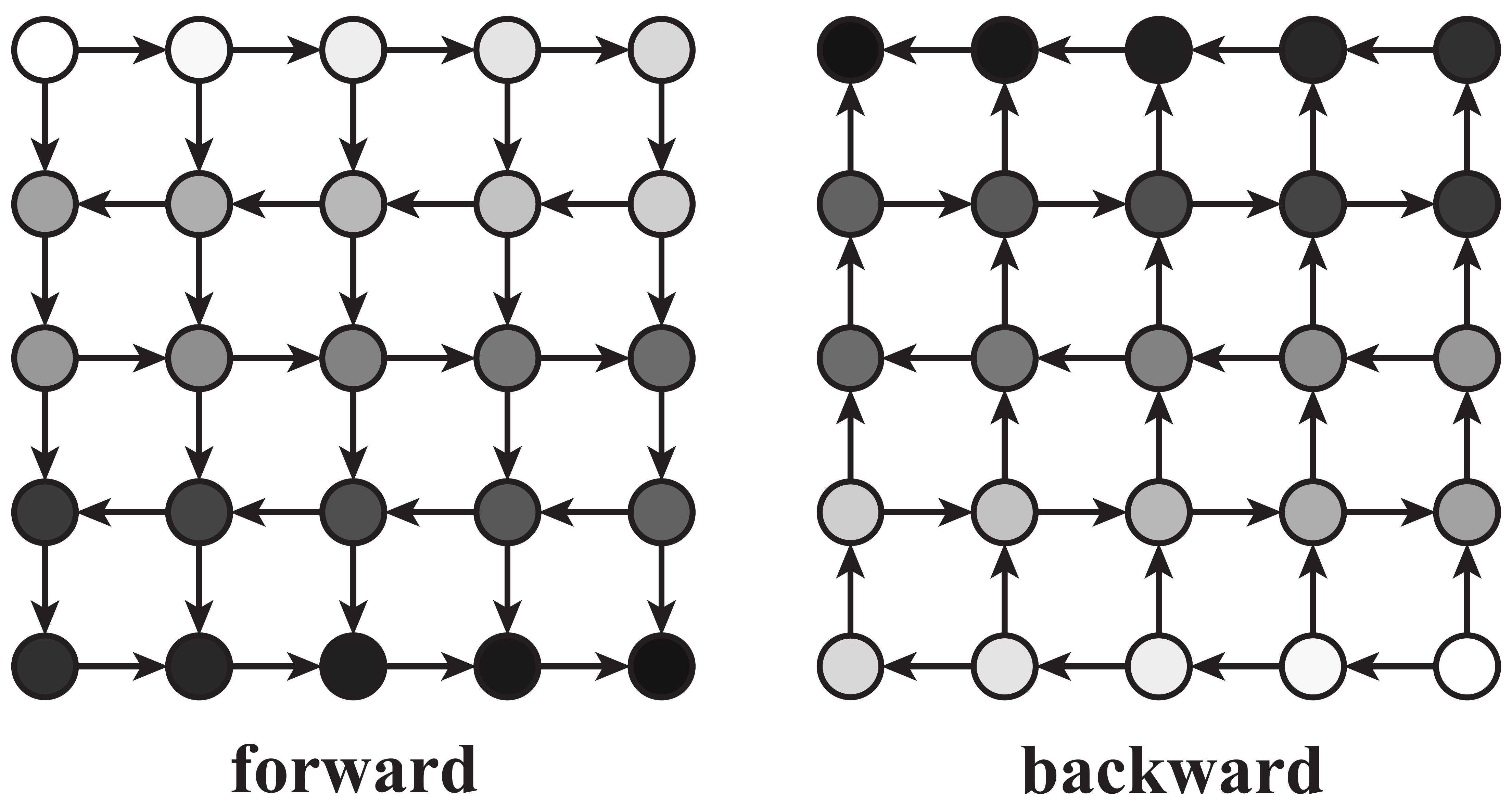}
              %\caption{}
             %\label{}
     \end{subfigure}
%\vspace{-3mm}
\caption{\small Coupled acyclic passes. 
We decouple an undirected cyclic graph into two directed acyclic graphs with each one allowing feed-forward computation. Inference is performed by alternately traversing the two acyclic graphs, while coupling their information at each step. 
%\textbf{Left} shows the information flow across multiple passes, while the red lines denotes the links used in updating a hidden state. 
%\textbf{Right} focuses on a particular pass. In updating a certain state, half of the context comes from the current pass, while the other half is from the previous pass.
}
%\vspace{-3mm}
\label{fig:graphrnn}
\end{figure}

\paragraph{\bf Coupled Acyclic Passes.}

In the proposed model, the dependencies among neighbors are \emph{cyclic}. 
Hence, the MAP estimate $\tilde{\vh} = f(\vx, \vtheta)$ cannot be computed in a single forward pass. 
Instead, Eq.\eqref{eq:h_rnn} needs to be applied across the graph in multiple iterations. 
Our strategy is to unroll this iterative inference procedure into multiple feed-forward passes along opposite directions, 
such that these passes together provide a complete context to each local estimate. 

Specifically, we decompose the underlying dependency graph $G = (V, E)$, which is undirected,
into two \emph{acyclic directed graphs} $G^f = (V, E^f)$ and $G^b = (V, E^b)$, as illustrated in Figure~\ref{fig:graphrnn}, 
such that each undirected edge $\{u, v\} \in E$ corresponds uniquely to an edge $(u, v) \in E^f$ 
and an opposite edge $(v, u) \in E^b$. 
It can be proved that such a decomposition always exists and 
that for each node $u \in V$, the neighborhood $\Nb_u$ can be expressed as
$\Nb_u = \Nb^f(u) \cup \Nb^b(u)$, where $\Nb^f(u)$ and $\Nb^b(u)$ are the set of parents of $u$ 
respectively along $G^f$ and $G^b$. 
%We provide a proof to this statement in the supplemental material.

Given such a decomposition, we can derive an iterative computational procedure, where each cycle couples
a \emph{forward pass} that applies Eq.\eqref{eq:h_rnn} along $G^f$ and a \emph{backward pass}\footnote{The word \emph{forward} and \emph{backward} here means the sequential order in the graph. They are not \emph{feed-forward} and \emph{back-propagation} in the context of deep neural networks.} along $G^b$.
After the $t$-th cycle, the state $h_u$ is updated to 
\begin{equation} \label{eq:cas}
h_u^{(t)} = \sigma \Bigg( 
	\sum_{v \in \Nb^f(u)} \left( \mW h_{v}^{(t-1)} + \mR x_{v} \right)  + 
	\sum_{v \in \Nb^b(u)} \left( \mW h_{v}^{(t)}   + \mR x_{v} \right) 
\Bigg).
\end{equation}
As states above, we have $\Nb_u = \Nb^f(u) \cup \Nb^b(u)$. 
Therefore, over a cycle, the updated state $h_u$ would incorporate information from all its neighbors. 
Note that a given graph $G$ can be decomposed in many different ways. 
In this work, we specifically choose the one that forms the \emph{zigzag} path. The advantage over a simple raster line order is that \emph{zigzag} path traverses all the nodes continuously, so that it conserves spatial coherence by making dependence of each node to all the previous nodes that have been visited before.
The forward and backward passes resulted from such decomposition are shown in Figure~\ref{fig:graphrnn}. 

This algorithm has two important properties:
First, the acyclic decomposition allows feed-forward computation as in Eq.\eqref{eq:h_rnn} to be applied. 
As a result, the entire inference procedure can be viewed as a feed-forward network that couples multiple RNNs operating along different directions.
Therefore, it can be learned in a way similar to other deep neural networks, using 
\textit{Stochastic Gradient Descent (SGD)}.
%or its variants~\cite{duchi2011adaptive,graves2013generating}.
%
Second, the feedback mechanism embodied by the backward pass facilitates the propagation of local information and thus the learning of long-range dependencies. 

%\paragraph{\bf Learning and Sampling Details.}

\paragraph{\bf Discussions with 2D-RNN.}
Previous work has explored two-dimensional extensions of RNN~\cite{mdrnn}, often referred to as \emph{2D-RNN}.
Such extensions, however, are formulated upon an acyclic graph, and 
can be considered as a trimmed down version of our algorithm. 
A major drawback of 2D-RNN is that it scans the image in a raster line order and it is not able to provide a feedback path. Therefore, the inference of each hidden state can only take into account $1/4$ of the context,
and there is no way to recover from a poor inference. 
As we will show in our experiments, this may cause undesirable effects.
Whereas bidirectional RNNs \cite{schuster1997bidirectional} may partly mitigate this problem,
they decouple the hidden states into multiple ones that are independent apriori, which would lead to consistency issues.
Recent work~\cite{berglund2015bidirectional} also finds it difficult to use in generative modeling.

\paragraph{\bf Implementation Details}
%For all our experiments, we actually use stacks of fully connected layers for modeling the hidden states potentials, as architectures of multiple layers prove to be more powerful in the deep learning community. 
For inference and learning, to make the computation feasible, we just take one forward pass and one backward pass. Thus, each node is only updated twice while being able to use the information from all possible contexts. The training patch size varies from 15 to 25 depending on the specific experiment. Overall, if we unroll the full inference procedure, our model\footnote{code available at https://github.com/zhirongw/deep-mrf} is more than thousands of layers deep. We use \textit{rmsprop}~\cite{graves2013generating} for optimization and we don't use dropout for regularization, as we find it oscillates the training. 
%Training our model generally takes about one day and 8GB memory of GPU. Inference takes several minutes depending on the image size.

\section{Experiments}

In the following experiments, we test the proposed deep MRF on 3 scenarios for modeling natural images. We first study its basic properties on \emph{texture synthesis}, and then we apply it on a prediction problem, \emph{image super-resolution}. Finally, we integrate global CNN models with local deep MRF for \emph{natural image synthesis}.

%In the first experiment, the model synthesizes stochastic textures by learning the pixel dependency in the spatial domain. In the second experiment, the model learns to generate high resolution images given low res- olution images. In the third experiment, we show how we can integrate global models, e.g. CNNs, for synthesizing images with scene structures and objects.

\subsection{Texture Synthesis}

\begin{figure}[t]
\begin{tabular}{ccccccc}
\centering
 & D12 & D34 & D104 & flowers & bark & clouds \\
\vspace{1pt} \parbox[t]{2mm}{\rotatebox[origin=c]{90}{input}}  &
\includegraphics[valign=m, width=0.15\textwidth]{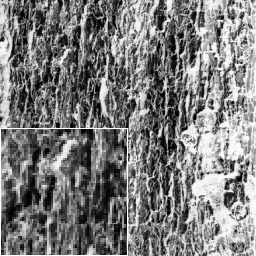}&
\includegraphics[valign=m, width=0.15\textwidth]{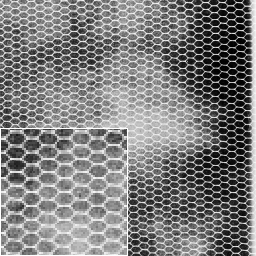} &
\includegraphics[valign=m, width=0.15\textwidth]{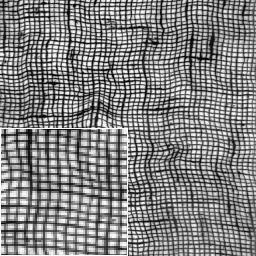} &
\includegraphics[valign=m, width=0.15\textwidth]{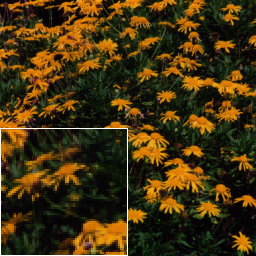} &
\includegraphics[valign=m, width=0.15\textwidth]{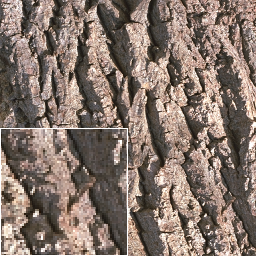} &
\includegraphics[valign=m, width=0.15\textwidth]{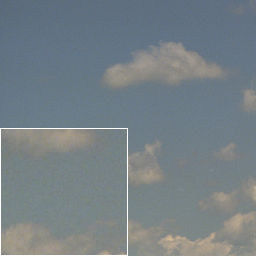}
\\
\\[-11pt]
 \parbox[t]{2mm}{\rotatebox[origin=c]{90}{2DRNN\cite{theis2015generative}}}  &
\includegraphics[valign=m, width=0.15\textwidth]{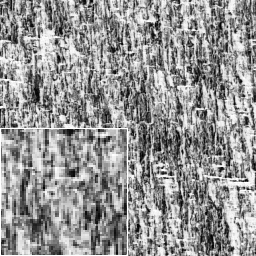} &
\includegraphics[valign=m, width=0.15\textwidth]{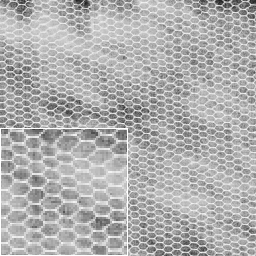} &
\includegraphics[valign=m, width=0.15\textwidth]{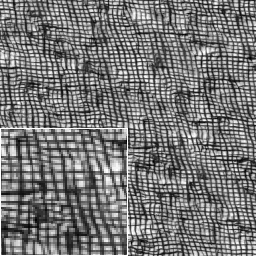} &
\includegraphics[valign=m, width=0.15\textwidth]{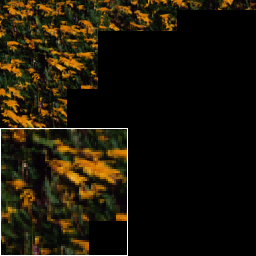} &
\includegraphics[valign=m, width=0.15\textwidth]{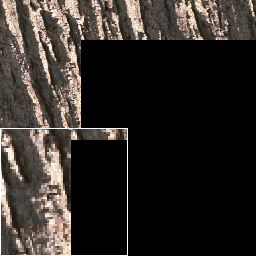} &
\includegraphics[valign=m, width=0.15\textwidth]{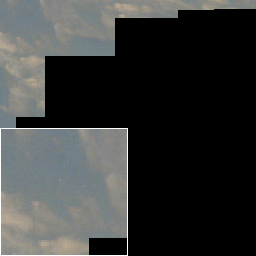}
\\
\\[-10pt]
 \parbox[t]{2mm}{\rotatebox[origin=c]{90}{graphics\cite{efros1999texture}}}  &
\includegraphics[valign=m, width=0.15\textwidth]{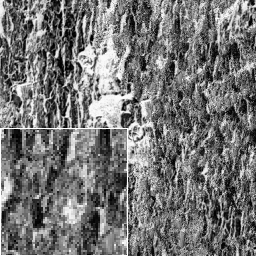} &
\includegraphics[valign=m, width=0.15\textwidth]{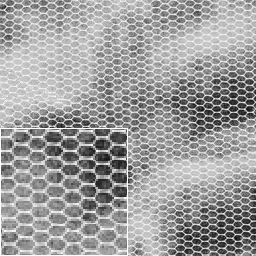} &
\includegraphics[valign=m, width=0.15\textwidth]{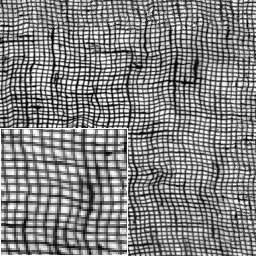} &
\includegraphics[valign=m, width=0.15\textwidth]{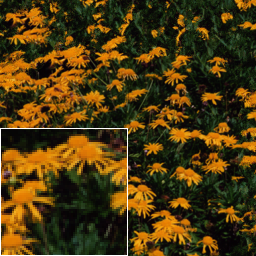} &
\includegraphics[valign=m, width=0.15\textwidth]{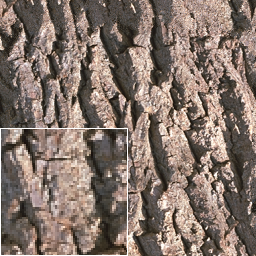} &
\includegraphics[valign=m, width=0.15\textwidth]{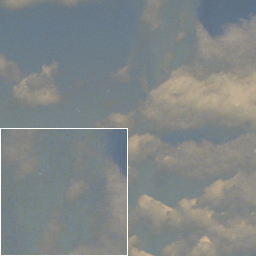}
\\
\\[-10pt]
\parbox[t]{2mm}{\rotatebox[origin=c]{90}{ours}}  &
\includegraphics[valign=m, width=0.15\textwidth]{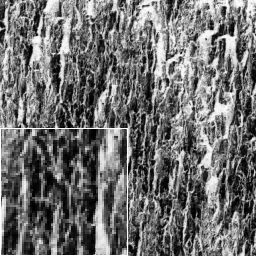} &
\includegraphics[valign=m, width=0.15\textwidth]{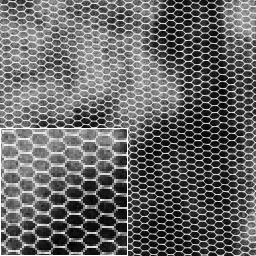} &
\includegraphics[valign=m, width=0.15\textwidth]{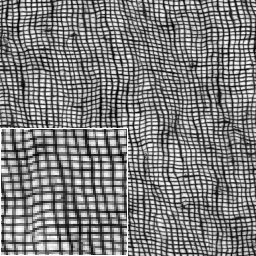} &
\includegraphics[valign=m, width=0.15\textwidth]{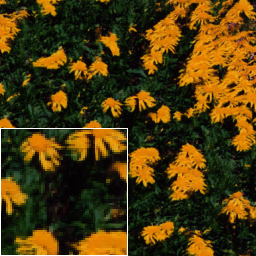} &
\includegraphics[valign=m, width=0.15\textwidth]{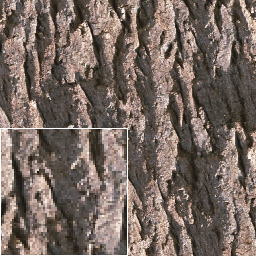} &
\includegraphics[valign=m, width=0.15\textwidth]{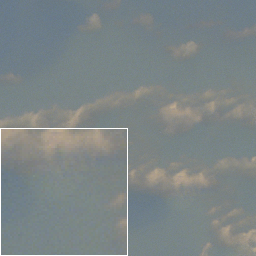}
\end{tabular}
\caption{Texture synthesis results.  }
%\vspace{-4mm}
\label{fig:texture}
\end{figure}

% briefly introduce the task first.
The task of texture synthesis is to synthesize new texture images that possess similar patterns and statistical characteristics as a given texture sample. 
The study of this problem originated from graphics~\cite{efros1999texture,wei2000fast}. 
The key to successful texture reproduction, as we learned from previous work, is to effectively capture the local patterns and variations.
Therefore, this task is a perfect testbed to assess a model's capability of modeling visual patterns. 

% describe the model settings
Our model works in a purely generative way. 
Given a sample texture, we train the model on randomly extracted patches of size $25 \times 25$, which are larger than most \emph{texels} in natural images.
We set $K = 20$, initialize $\vx$ and $\vh$ to zeros, and train the model with back-propagation along the coupled acyclic graph. 
%We use 20 mixtures of gaussians for the output. 
With a trained model, we can generate textures by running the RNN to derive the latent states and at the same time sampling the output pixels. 
As our model is stationary, it can generate texture images of arbitrary sizes.

We work on two texture datasets, Brodatz~\cite{brodatz} for grayscale images, and VisTex~\cite{vistex} for color images. From the results shown in Figure~\ref{fig:texture}, our synthesis visually resembles to high resolution natural images, and the quality is close to the non-parametric approach~\cite{efros1999texture}. 
We also compare with the 2D-RNN.~\cite{theis2015generative}. As we can see, the results obtained using 2D-RNN, which synthesizes based only on the left and upper regions, exhibit undesirable effects and often evolve into blacks in the bottom-right parts. 

\begin{figure}[t]
\begin{tabular}{ccccccc}
\Xhline{2\arrayrulewidth}
 & 5 & 10 & 15 & 20 & 25 & input
\\
\multirow{2}{*}{\parbox[t]{2mm}{\rotatebox[origin=c]{90}{training patch size}}}
&
\includegraphics[valign=m, width=0.15\textwidth]{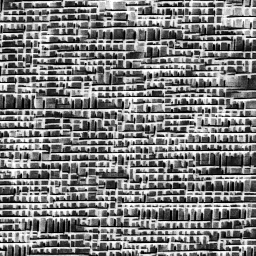}&
\includegraphics[valign=m, width=0.15\textwidth]{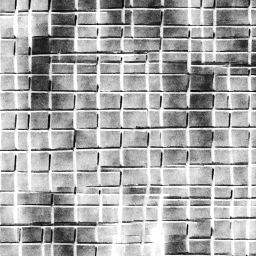}&
\includegraphics[valign=m, width=0.15\textwidth]{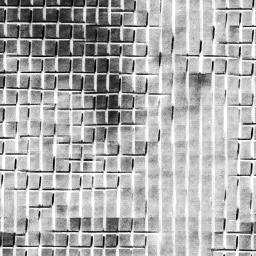}&
\includegraphics[valign=m, width=0.15\textwidth]{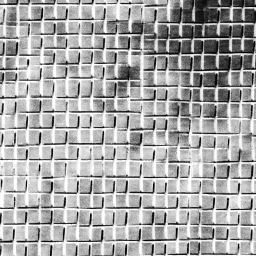}&
\includegraphics[valign=m, width=0.15\textwidth]{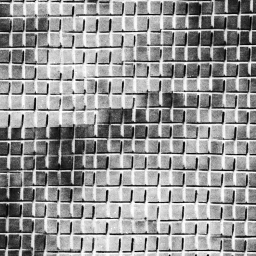} & 
\includegraphics[valign=m, width=0.15\textwidth]{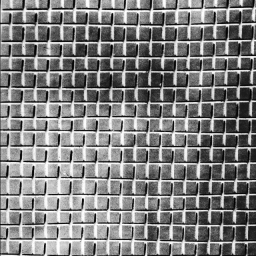}
\\
\\[-10pt]
 &
\includegraphics[valign=m, width=0.15\textwidth]{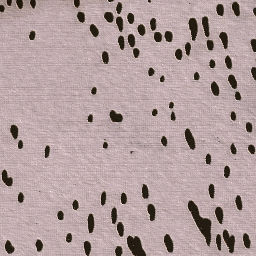}&
\includegraphics[valign=m, width=0.15\textwidth]{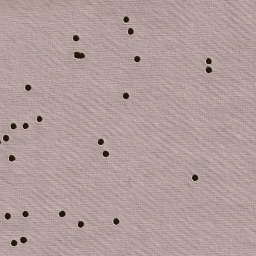}&
\includegraphics[valign=m, width=0.15\textwidth]{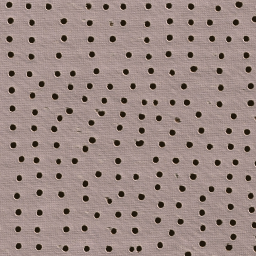}&
\includegraphics[valign=m, width=0.15\textwidth]{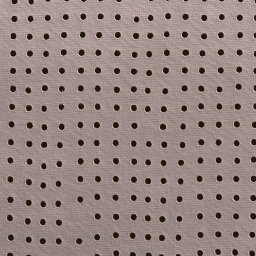}&
\includegraphics[valign=m, width=0.15\textwidth]{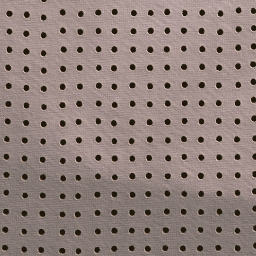}&
\includegraphics[valign=m, width=0.15\textwidth]{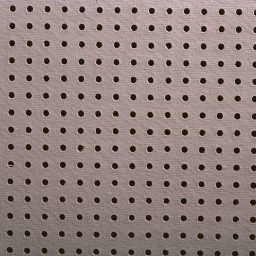}
\\
\\[-10pt]
\Xhline{2\arrayrulewidth}
 & 1 & 2 & 5 & 10 & 20 & input
\\
\multirow{2}{*}{\parbox[t]{2mm}{\rotatebox[origin=c]{90}{number of mixtures}}}
&
\includegraphics[valign=m, width=0.15\textwidth]{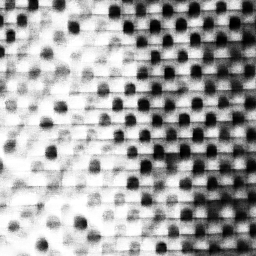}&
\includegraphics[valign=m, width=0.15\textwidth]{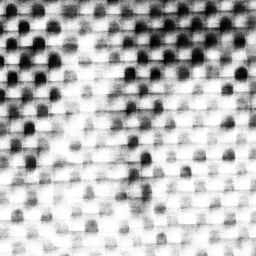}&
\includegraphics[valign=m, width=0.15\textwidth]{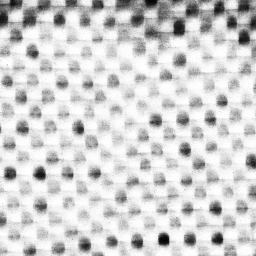}&
\includegraphics[valign=m, width=0.15\textwidth]{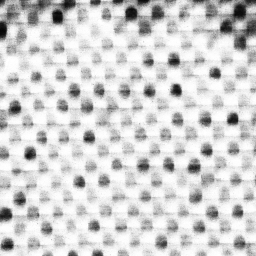}&
\includegraphics[valign=m, width=0.15\textwidth]{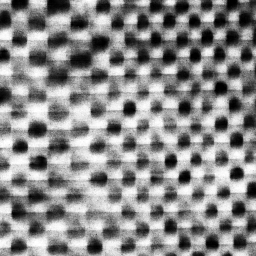}&
\includegraphics[valign=m, width=0.15\textwidth]{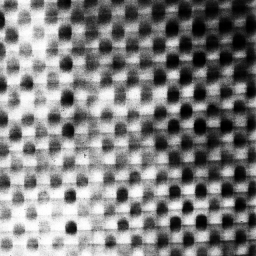}
\\
\\[-10pt]
&
\includegraphics[valign=m, width=0.15\textwidth]{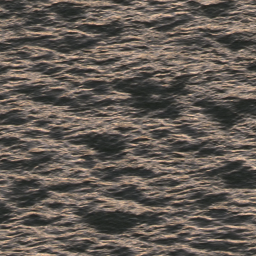}&
\includegraphics[valign=m, width=0.15\textwidth]{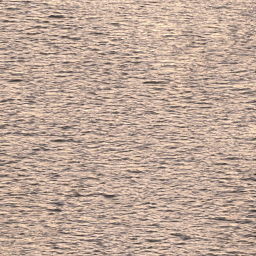}&
\includegraphics[valign=m, width=0.15\textwidth]{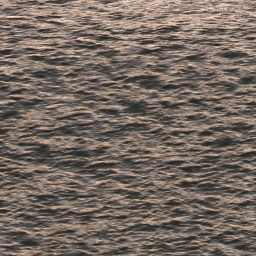}&
\includegraphics[valign=m, width=0.15\textwidth]{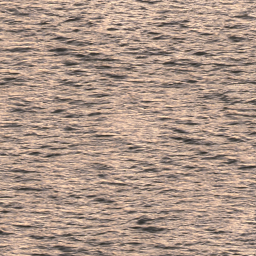}&
\includegraphics[valign=m, width=0.15\textwidth]{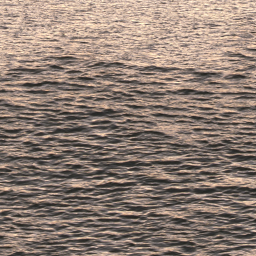}&
\includegraphics[valign=m, width=0.15\textwidth]{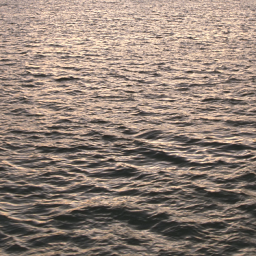}
\end{tabular}
\caption{Texture synthesis by varying the patch size and the number of mixtures. }
%\vspace{-2mm}
\label{fig:texture2}
\end{figure}

% analysis
Two fundamental parameters control the behaviors of our texture model. The training patch size decides the farthest spatial relationships that could be learned from data. The number of gaussian mixtures control the dynamics of the texture landscape. We analyze our model by changing the two parameters. As shown in Figure~\ref{fig:texture2}, bigger training patch size and bigger number of mixtures consistently improves the results. For non-parametric approaches, bigger patch size would dramatically bring up the computation cost. While for our model, the inference time holds the same regardless of the patch size that the model is trained on.  Moreover, our parametric model is able to scale to large dataset without bringing additional computations.

%When synthesizing a 256x256 image, our parametric model is orders of faster (2 mins compared with 1 hour) while being able to scale to larger dataset.

\begin{figure}[t]
\centering
     \begin{subfigure}{0.30\textwidth}
			\centering
             \includegraphics[width=1\textwidth]{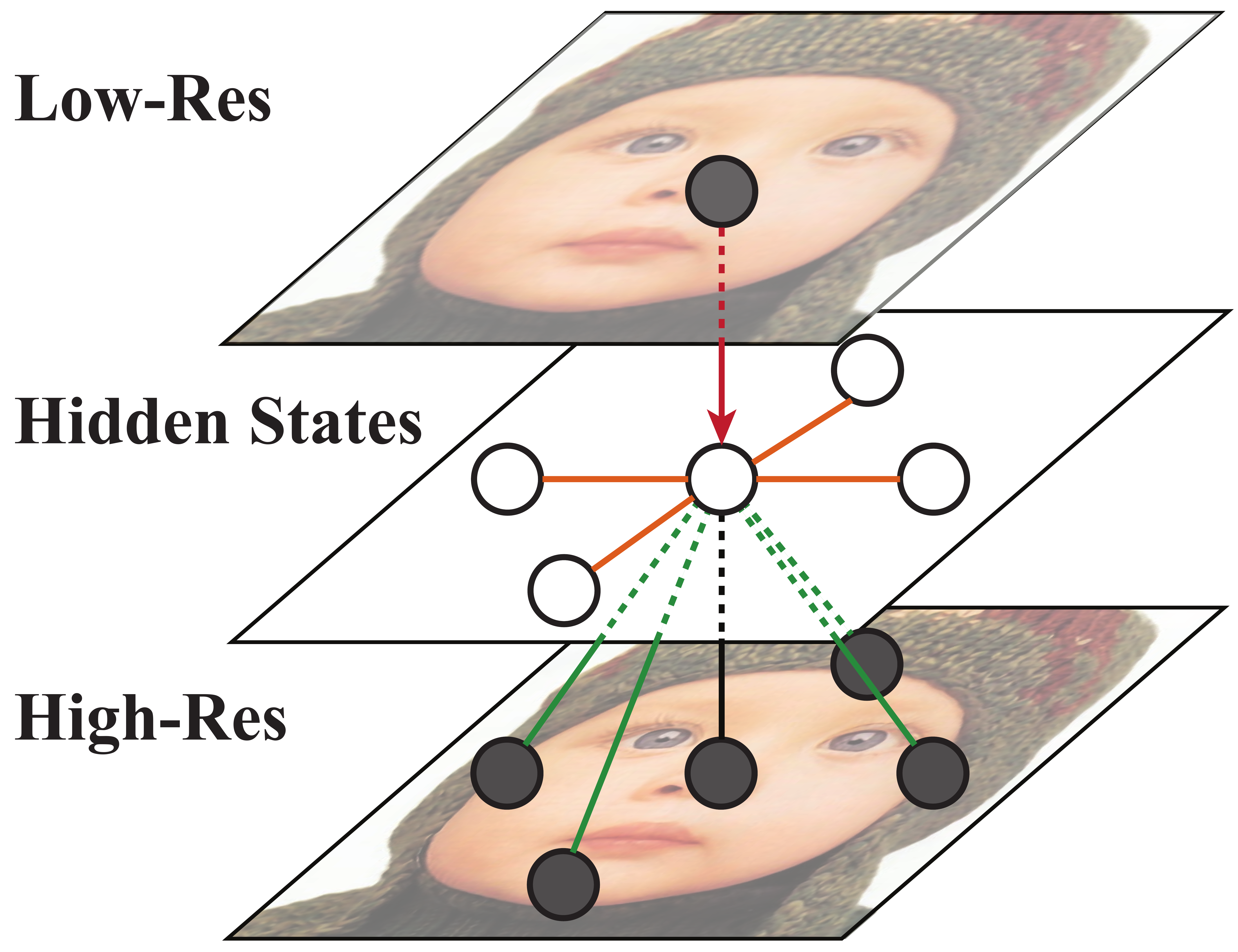}
              %\caption{}
             %\label{}
     \end{subfigure}
     ~
     	\begin{subfigure}{0.65\textwidth}
			\centering
             \includegraphics[width=1\textwidth]{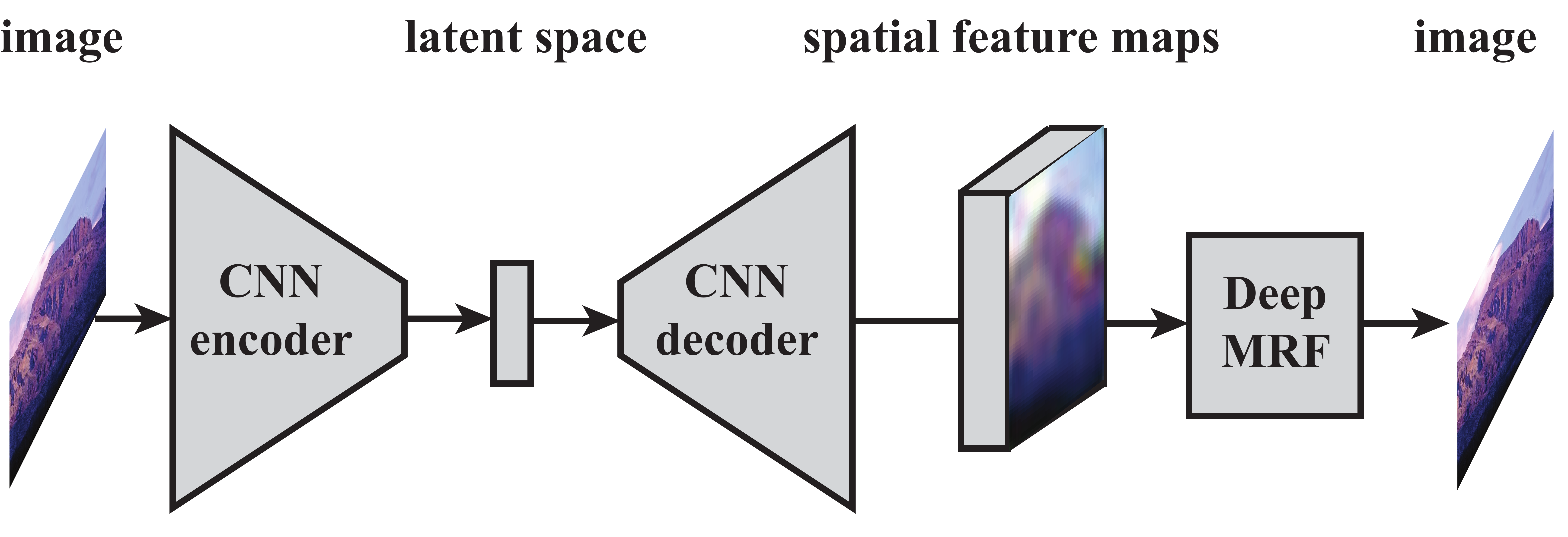}
              %\caption{}
              %\label{}
     \end{subfigure}
%\vspace{-3mm}
\caption{Adapting deep MRFs to specific applications. Image super-resolution: the hidden state receives an additional connection from the low-resolution pixel. Image synthesis: deep MRF renders the final image from a spatial feature map, which is jointly learned by a variational auto-encoder.}
%\vspace{-4mm}
\label{fig:exp_pipeline}
\end{figure}

\begin{table}[t]
\centering
\caption{PSNR (dB) on Set5 dataset with upscale factor 2,3,4.}
\footnotesize
\begin{adjustbox}{width=1\textwidth}
\begin{tabular}{c|c|c|c|c|c|c|c|c|c|c|c|c}
\Xhline{2\arrayrulewidth}
\multirow{2}{*}{images} & \multicolumn{4}{c|}{2x upscale} & \multicolumn{4}{c|}{3x upscale}  & \multicolumn{4}{c}{4x upscale}  \\
\cline{2-13}
  & Bicubic &  CNN & SE & Ours  & Bicubic & CNN & SE & Ours  & Bicubic  & CNN & SE & Ours \\
\hline
  %\Xhline{2\arrayrulewidth}
 baby & 37.07 & 38.30 & \textbf{38.48} & 38.31 & 33.91 & 35.01&  \textbf{35.22} & 35.15 &  31.78  & 32.98 & \textbf{33.14} & 32.94 \\
\hline
 bird & 36.81 & 40.40 & \textbf{40.50}  & 40.36 & 32.58 & 34.91 & 35.58 &  \textbf{36.14} & 30.18 &  31.98 & \textbf{32.54} & 32.49\\
\hline
% \tabularnewline
 butterfly & 27.43 &  32.20 & 31.86 & \textbf{32.74} & 24.04  & 27.58 & 26.86 &  \textbf{29.09} & 22.10 & 25.07 & 24.09 &  \textbf{25.78}\\
\hline 
% \tabularnewline
 head & 34.86 &  35.64 & 35.69 & \textbf{35.70} & 32.88  & 33.55 & \textbf{33.76} &  33.63 & 31.59 & 32.19 & \textbf{32.52} & 32.41\\
\hline 
women & 32.14 &  34.94 & \textbf{35.33}  & 34.84 & 28.56 & 30.92 & 31.36 & \textbf{31.69} & 26.46 & 28.21 & 28.92 &  \textbf{28.97} \\
\Xhline{2\arrayrulewidth}
average & 33.66  & 36.34 & 36.37 & \textbf{36.38} & 31.92 & 32.30 & 32.56 & \textbf{33.14} & 28.42 & 30.09 &30.24 & \textbf{30.52}\\
\Xhline{2\arrayrulewidth}
\end{tabular}
\end{adjustbox}
%\vspace{-2mm}
\label{table:sr5}
\end{table}

\begin{table}[t]
\centering
\caption{PSNR (dB) on various dataset with upscale factor 3.}
\footnotesize
\begin{tabular}{c|c|c|c|c|c|c}
\Xhline{2\arrayrulewidth}
Dataset & Bicubic & A+~\cite{timofte2014a+} &  CNN~\cite{dong2014learning} & SE~\cite{huang2015single} & CSCN~\cite{wang2015deep} & Ours \\
%\hline
\Xhline{2\arrayrulewidth}
%\hline
Set5 & 30.39 &  32.59 & 32.30 & 32.56 & 33.10 & \textbf{33.14}\\
\hline
Set14 & 27.54 & 29.13 & 29.00 & 29.16 & \textbf{29.41} & 29.38\\
\hline
BSD100 &  27.22 & 28.18 & 28.20 & 28.20 & 28.50 & \textbf{28.54} \\
\Xhline{2\arrayrulewidth}
\end{tabular}
%\vspace{-2mm}
\label{table:sr_x3}
\end{table}

\iffalse
\begin{table}[t]
\centering
\caption{PSNR (dB) on Set14 dataset with upscale factor 3.}
\footnotesize
\begin{tabular}{c|c|c|c|c|c|c|c|c}
\Xhline{2\arrayrulewidth}
   & baboon & barbara & bridge & guard & comic & face & flowers \\
\hline
bicubic & 23.21 & 26.25 & 24.40 & 26.55 & 23.12 & 32.82 & 27.23 \\
\hline
ANR & 23.56 & 26.69 & 25.01 & 27.08 & 24.04 & 33.62 & 28.49 \\
\hline
CNN & 23.60 & 26.66 & 25.07 & 27.20 & 24.39 & 33.58 & 28.97 \\
\hline
Ours & \textbf{23.66} & \textbf{27.08} &  \textbf{25.38} & \textbf{27.35} & \textbf{24.9} & \textbf{33.71} & \textbf{29.58}\\
\hline
& foreman & lenna & man & monarch & pepper & ppt3 & zebra & average \\
\hline
bicubic  & 31.18 & 31.68 & 27.01 & 29.43 & 32.39 & 23.71 & 26.63 & 27.54 \\
\hline
ANR & 33.23 & 33.08 & 27.92 & 31.09 & 33.82 & 25.03 & 28.43 & 28.65 \\
\hline
CNN & 33.35 & 33.39 & 28.18 & 32.39 & 34.35 & 26.02 & 28.87 & 29.00 \\
\hline
Ours & \textbf{33.4} & \textbf{33.7} & \textbf{28.48} & \textbf{33.58} & \textbf{34.39} & \textbf{26.87} & \textbf{29.3} & \textbf{29.38}\\
\Xhline{2\arrayrulewidth}
\end{tabular}
%\vspace{-2mm}
\label{table:sr14}
\end{table}
\fi

\subsection{Image Super-Resolution}
% introduce the task
Image super-resolution is a task to produce a high resolution image given a single low resolution one. 
Whereas previous MRF-based models~\cite{freeman2000learning,freeman2011markov} work reasonably,
the quality of their products is inferior to the state-of-the-art models based on deep learning~\cite{dong2014learning,wang2015deep}. 
With deep MRF, we wish to close the gap.

% our approach
%
Unlike in texture synthesis, the generation of this task is driven by a low-resolution image. 
To incorporate this information, we introduce additional connections between the hidden states and corresponding pixels of the low-resolution image,
as shown in Figure~\ref{fig:exp_pipeline}.
It is noteworthy that we just input \emph{a single pixel} (instead of a \emph{patch}) at each site, and in this way, we can test whether the model can propagate information across the spatial domain.
As the task is deterministic, we use a GMM with a single component and fix its variance. 
In the testing stage, we output the mean of the Gaussian component at each location as the inferred high-resolution pixel. 
This approach is very generic -- the model is not specifically tuned for the task and no pre- and post-processing steps are needed. 

%receives an additional connection from the low resolution image. In this way, deep MRFs work like conditional random field. While we can actually feed patches from the low resolution image to each location state, to test whether our model can itself learn to propagate that information from the spatial domain, we just input \textit{a single pixel} at each spatial location. Since the task defined is deterministic, we use a single mixture for the GMM and fix the variance to be constant. During testing, we output the mean of the gaussian as the most probable sample. As we can see, our approach is very generic, with no modules specifically tuned for this problem. We also don't need any kind of pre-processing and post-processing, such as shaving the image borders. An illustration of our approach is shown in Figure~\ref{fig:exp_pipeline}.

% experimental settings
We train our model on a widely used super-resolution dataset~\cite{bevilacqua2012low} which contains $91$ images, and test it on Set5, Set14, and BSD100~\cite{MartinFTM01}. The training is on patches of size $16 \times 16$ and \emph{rmsprop} with momentum $0.95$ is used. 
We use PSNR for quantitative evaluation. Following previous work, we only consider the luminance channel in the \textit{YCrCb} color space. The two chrominance channels are upsampled with bicubic interpolation. 

\begin{figure}[t]
\scriptsize
\centering
\begin{tabular}{cccc}
\centering
\includegraphics[width=0.22\textwidth]{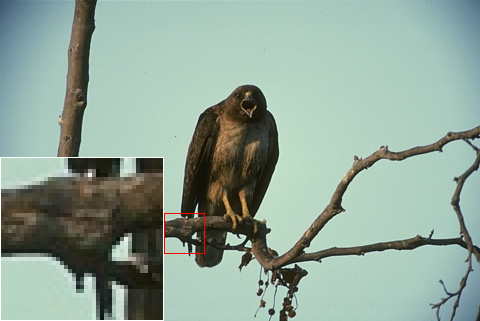} &
\includegraphics[width=0.22\textwidth]{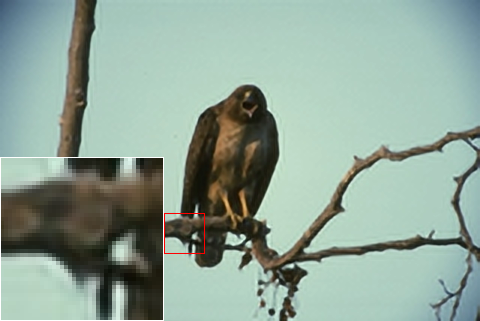} &
\includegraphics[width=0.22\textwidth]{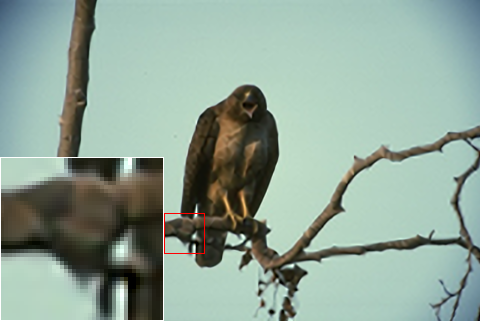} &
\includegraphics[width=0.22\textwidth]{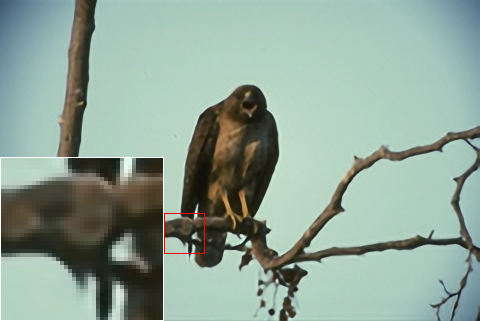}  
\\
\centering
Original  / PSNR &  CNN / 31.65 dB & SE / 31.56 dB & Ours / \textbf{33.11 dB} \\
%\centering
%\includegraphics[width=0.23\textwidth]{figures/SR/baby/12.pdf} &
%\includegraphics[width=0.23\textwidth]{figures/SR/baby/72.pdf} &
%\includegraphics[width=0.23\textwidth]{figures/SR/baby/82.pdf} &
%\includegraphics[width=0.23\textwidth]{figures/SR/baby/102.pdf} 
%\\
%\centering
%Original  / PSNR & ANR / 35.13 dB & CNN / 35.01 dB & Ours / \textbf{35.15 dB} \\
\centering
\includegraphics[width=0.22\textwidth]{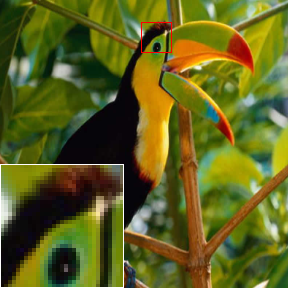} &
\includegraphics[width=0.22\textwidth]{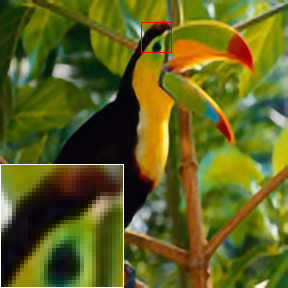} &
\includegraphics[width=0.22\textwidth]{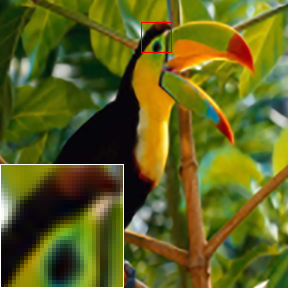} &
\includegraphics[width=0.22\textwidth]{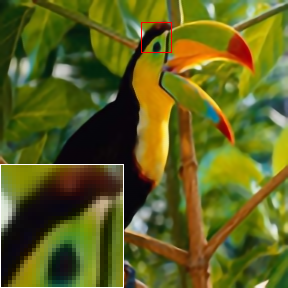}
\\
\centering
Original  / PSNR &  CNN / 34.91 dB & SE / 35.58 dB & Ours / \textbf{36.14 dB} \\
%\centering
%\includegraphics[width=0.23\textwidth]{figures/SR/head/12.pdf} &
%\includegraphics[width=0.23\textwidth]{figures/SR/head/72.pdf} &
%\includegraphics[width=0.23\textwidth]{figures/SR/head/82.pdf} &
%\includegraphics[width=0.23\textwidth]{figures/SR/head/102.pdf} 
%\\
%\centering
%Original  / PSNR & ANR / 33.63 dB & CNN / 33.55 dB & Ours / \textbf{33.63 dB} \\
\centering
\includegraphics[width=0.22\textwidth]{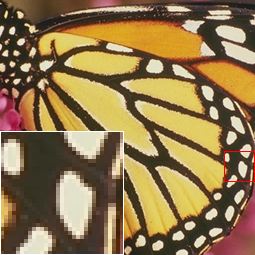} &
\includegraphics[width=0.22\textwidth]{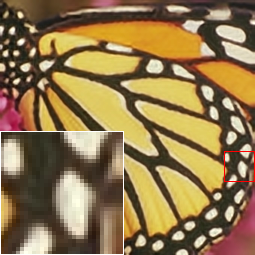} &
\includegraphics[width=0.22\textwidth]{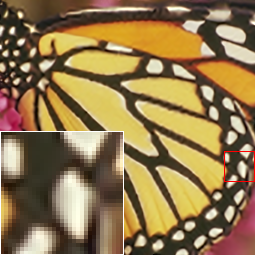} &
\includegraphics[width=0.22\textwidth]{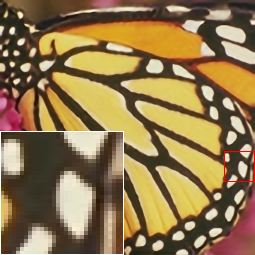} 
\\
\centering
Original  / PSNR &  CNN / 27.58 dB & SE / 26.86 dB & Ours / \textbf{29.09 dB} \\
\end{tabular}
\caption{Image super resolution results from Set 5 with upscaling factor 3.  }
%\vspace{-5mm}
\label{fig:sr}
\end{figure}

As shown in Table~\ref{table:sr5} and Table~\ref{table:sr_x3}, our approach outperforms the CNN-based baseline~\cite{dong2014learning} and compares favorably with the state-of-the-art methods dedicated to this task~\cite{huang2015single,wang2015deep}. One possible explanation for the success is that our model not only learns the mapping, but also learns the image statistics for high resolution images. The training procedure which unrolls the RNN into thousands of steps that share parameters also reduces the risk of overfitting. 
The results also demonstrate the particular strength of our model in handling large upscaling factors and difficult images.
Figure~\ref{fig:sr} shows several examples visually.

\begin{figure}[t]
\begin{tabular}{cccccccc|c|c}
\\
 & \multicolumn{7}{c|}{Deep MRF} & VAE & DCGAN
 \\
\\[-10pt]
\parbox[t]{2mm}{\rotatebox[origin=c]{90}{street}}  &
\includegraphics[valign=m, width=0.10\textwidth]{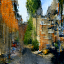}&
\includegraphics[valign=m, width=0.10\textwidth]{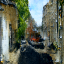} &
\includegraphics[valign=m, width=0.10\textwidth]{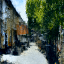} &
\includegraphics[valign=m, width=0.10\textwidth]{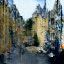} &
\includegraphics[valign=m, width=0.10\textwidth]{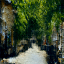} &
\includegraphics[valign=m, width=0.10\textwidth]{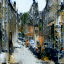} &
\includegraphics[valign=m, width=0.10\textwidth]{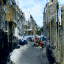} &
\includegraphics[valign=m, width=0.10\textwidth]{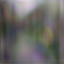} &
\includegraphics[valign=m, width=0.10\textwidth]{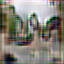}
\\
\\[-10pt]
\parbox[t]{2mm}{\rotatebox[origin=c]{90}{river}}  &
\includegraphics[valign=m, width=0.10\textwidth]{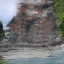}&
\includegraphics[valign=m, width=0.10\textwidth]{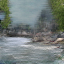} &
\includegraphics[valign=m, width=0.10\textwidth]{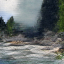} &
\includegraphics[valign=m, width=0.10\textwidth]{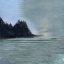} &
\includegraphics[valign=m, width=0.10\textwidth]{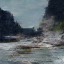} &
\includegraphics[valign=m, width=0.10\textwidth]{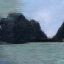} &
\includegraphics[valign=m, width=0.10\textwidth]{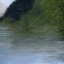} &
\includegraphics[valign=m, width=0.10\textwidth]{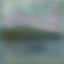} &
\includegraphics[valign=m, width=0.10\textwidth]{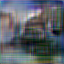}
\\
\\[-10pt]
\parbox[t]{2mm}{\rotatebox[origin=c]{90}{barn}}  &
\includegraphics[valign=m, width=0.10\textwidth]{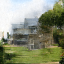}&
\includegraphics[valign=m, width=0.10\textwidth]{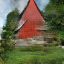} &
\includegraphics[valign=m, width=0.10\textwidth]{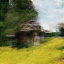} &
\includegraphics[valign=m, width=0.10\textwidth]{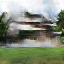} &
\includegraphics[valign=m, width=0.10\textwidth]{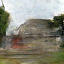} &
\includegraphics[valign=m, width=0.10\textwidth]{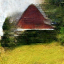} &
\includegraphics[valign=m, width=0.10\textwidth]{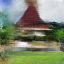} &
\includegraphics[valign=m, width=0.10\textwidth]{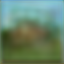} &
\includegraphics[valign=m, width=0.10\textwidth]{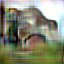}
\\
\iffalse
\vspace{1pt} \parbox[t]{2mm}{\rotatebox[origin=c]{90}{valley1}}  &
\includegraphics[valign=m, width=0.10\textwidth]{figures/G/Our/valley/6.png}&
\includegraphics[valign=m, width=0.10\textwidth]{figures/G/Our/valley/7.png} &
\includegraphics[valign=m, width=0.10\textwidth]{figures/G/Our/valley/9.png} &
\includegraphics[valign=m, width=0.10\textwidth]{figures/G/Our/valley/10.png} &
\includegraphics[valign=m, width=0.10\textwidth]{figures/G/Our/valley/19.png} &
\includegraphics[valign=m, width=0.10\textwidth]{figures/G/Our/valley/21.png} &
\includegraphics[valign=m, width=0.10\textwidth]{figures/G/Our/valley/22.png} &
\includegraphics[valign=m, width=0.10\textwidth]{figures/G/VAE/valley/2_gen.png} &
\includegraphics[valign=m, width=0.10\textwidth]{figures/G/DCGAN/valley/2.png}
\\
\fi
\\[-10pt]
\parbox[t]{2mm}{\rotatebox[origin=c]{90}{valley}}  &
\includegraphics[valign=m, width=0.10\textwidth]{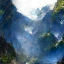}&
\includegraphics[valign=m, width=0.10\textwidth]{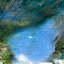} &
\includegraphics[valign=m, width=0.10\textwidth]{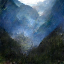} &
\includegraphics[valign=m, width=0.10\textwidth]{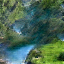} &
\includegraphics[valign=m, width=0.10\textwidth]{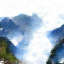} &
\includegraphics[valign=m, width=0.10\textwidth]{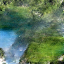} &
\includegraphics[valign=m, width=0.10\textwidth]{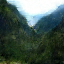} &
\includegraphics[valign=m, width=0.10\textwidth]{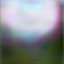} &
\includegraphics[valign=m, width=0.10\textwidth]{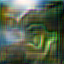}
\\
%\\[-10pt]
\parbox[t]{2mm}{\rotatebox[origin=c]{90}{mountain}}  &
\includegraphics[valign=m, width=0.10\textwidth]{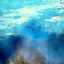}&
\includegraphics[valign=m, width=0.10\textwidth]{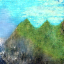} &
\includegraphics[valign=m, width=0.10\textwidth]{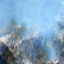} &
\includegraphics[valign=m, width=0.10\textwidth]{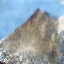} &
\includegraphics[valign=m, width=0.10\textwidth]{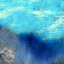} &
\includegraphics[valign=m, width=0.10\textwidth]{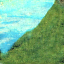} &
\includegraphics[valign=m, width=0.10\textwidth]{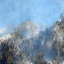} &
\includegraphics[valign=m, width=0.10\textwidth]{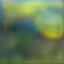} &
\includegraphics[valign=m, width=0.10\textwidth]{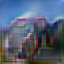}
\\
\\[-10pt]
\parbox[t]{2mm}{\rotatebox[origin=c]{90}{alley}}  &
\includegraphics[valign=m, width=0.10\textwidth]{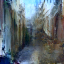}&
\includegraphics[valign=m, width=0.10\textwidth]{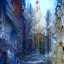} &
\includegraphics[valign=m, width=0.10\textwidth]{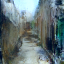} &
\includegraphics[valign=m, width=0.10\textwidth]{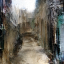} &
\includegraphics[valign=m, width=0.10\textwidth]{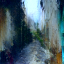} &
\includegraphics[valign=m, width=0.10\textwidth]{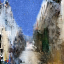} &
\includegraphics[valign=m, width=0.10\textwidth]{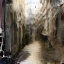} &
\includegraphics[valign=m, width=0.10\textwidth]{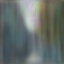} &
\includegraphics[valign=m, width=0.10\textwidth]{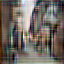}
\\
\end{tabular}
%\vspace{-3mm}
\caption{Image synthesis results.  }
%\vspace{-4mm}
\label{fig:synthesis}
\end{figure}

%\vspace{-3mm}
\subsection{Natural Image Synthesis}
% problem intro
Images can be roughly considered as a composition of textures with the guidance of scene and object structures. 
In this task, we move beyond the synthesis of homogeneous textures, and try to generate natural images with structural guidance.

% our technical overview
While our model excels in capturing spatial dependencies, learning weak dependencies across the entire image is both computationally infeasible and analytically inefficient. Instead, we adopt a global model to capture the overall structure and use it to provide contextual guidance to MRF.
Specifically, we incorporate the \emph{variational auto-encoder (VAE)}~\cite{kingma2013auto} for this purpose -- 
VAE generates feature maps at each location and our model uses that feature to render the final image (see Figure~\ref{fig:exp_pipeline}). 
Such features may contain information of scene layouts, objects, and texture categories.

% technical details
We train the joint model end-to-end from scratch. During each iteration, the VAE first encodes the image into a latent vector, then decodes it to a feature map with the same size of the input image. We then connect this feature map to the latent states of the deep MRF.
The total loss is defined as the addition of gaussian mixtures at image space and KL divergence at high-level VAE latent space. For training, we randomly extracts patches from the feature map. The gradients from the deep MRF back to the VAE thus only cover the patches being extracted. During testing, VAE randomly samples from the latent space and decodes it to generate the global feature maps.  The output pixels are sampled from the GMM with 10 mixtures along the coupled acyclic graph. 
%Overall, the computation pipeline is similar to the image super-resolution experiment, except that the conditional feature information is learned.

% experimental settings
We work on the MSRC~\cite{shotton2009textonboost} and SUN database~\cite{xiao2010sun} and select some scene categories with rich natural textures, such as \emph{Mountains} and \emph{Valleys}. 
Each category contains about a hundred images. As we will see, our approach generalizes much better than the data-hungry CNN approaches. 
We train the model on images of size $64 \times 64$ with a batch size of $4$. 
For each image, we extract $16$ patches of size $15 \times 15$ for training. 
Figure~\ref{fig:synthesis} shows several images generated from our models, in comparison with those obtained from the baselines, namely raw VAE~\cite{kingma2013auto} and DCGAN~\cite{radford2015unsupervised}. The CNN architecture is shared for all methods described in the DCGAN paper~\cite{radford2015unsupervised} to ensure fair comparison. 
We can see our model successfully captures a variety of local patterns, such as water, clouds, wall and trees. The global appearance also looks coherent, real and dynamic. The state-of-the-art CNN based models, which focuses too much on global structures, often yield sub-optimal local effects.

%\vspace{-2mm}
\section{Conclusions}
%\vspace{-4mm}
% overall contribution
We present a new class of MRF model whose potential functions are expressed by powerful fully-connected neurons.
% theoretical contribution, technical contribution
Through theoretical analysis, we draw close connections between probabilistic deep MRFs and end-to-end RNNs.
To tackle the difficulty of inference in cyclic graphs, we derive a new framework that decouples a cyclic graph with multiple coupled acyclic passes.
% experimental results
Experimental results show state-of-the-art results on a variety of low-level vision problems, which demonstrate the strong capability of MRFs with expressive potential functions.

\paragraph{\textnormal{\textbf{Acknowledgment.}}}
This work is supported by the Big Data Collaboration Research grant (CUHK Agreement No. TS1610626) and the Early Career Scheme (ECS) grant (No: 24204215). We also thank Aditya Khosla for helpful discussions and comments on a draft of the manuscript.

\clearpage

\bibliographystyle{splncs}
\bibliography{pixel_refs}
\end{document}